\documentclass[final,12pt]{colt2019}

\usepackage{times}

\usepackage{amsmath}
\usepackage{amssymb}
\usepackage{mathrsfs}
\usepackage{bbm}
\usepackage{bm}
\usepackage{multirow}
\usepackage{enumitem}
\usepackage{marginnote}
\usepackage{graphicx}
\usepackage{mdframed}
\usepackage{mathcomp}
\usepackage{stmaryrd}
\usepackage{mathtools}
\usepackage{xr}
\usepackage{nicefrac}       
\usepackage{microtype}      

\DeclareMathOperator{\T}{\intercal}

\DeclareMathOperator{\Grad}{grad}

\DeclareMathOperator*{\argmin}{argmin}

\newcommand*{\bh}[1]{\widehat{\bm{#1}}}
\newcommand*{\what}[1]{\widehat{#1}}

\newcommand*{\ceil}[1]{\lceil{#1}\rceil}

\newcommand*{\op}[1]{\operatorname{#1}}

\newcommand{\inner}[2]{\langle#1,#2\rangle}

\newcommand{\norm}[1]{\left\lVert#1\right\rVert}

\newcommand{\grad}[1]{\Grad #1}

\newcommand{\tmp}[1]{#1}

\newmdtheoremenv{condition}{Condition}

\newcommand{\vertiii}[1]{{\left\vert\kern-0.25ex\left\vert\kern-0.25ex\left\vert #1 
    \right\vert\kern-0.25ex\right\vert\kern-0.25ex\right\vert}}

\makeatletter
\newcommand*\bcdot{\mathpalette\bigcdot@{.5}}
\newcommand*\bigcdot@[2]{\mathbin{\vcenter{\hbox{\scalebox{#2}{$\m@th#1\bullet$}}}}}

\newlength\myindent
\usepackage{algorithm}
\usepackage{algorithmic}
\usepackage{comment}

\usepackage{todonotes}
\usepackage{bbding}
\usepackage{booktabs}
\usepackage{footnote}
\makesavenoteenv{tabular}
\makesavenoteenv{table}
\let\norm\undefined
\let\set\undefined

\let\ceil\undefined
\usepackage{notation}
\usepackage{bm}
\usetikzlibrary{patterns,snakes}

\newcommand{\loss}{l}
\DeclareBoldMathCommand{\vloss}{l}
\DeclareBoldMathCommand{\grad}{g}
\DeclareBoldMathCommand{\fakegrad}{\mathring{\bm{g}}}
\DeclareBoldMathCommand{\e}{e}
\DeclareBoldMathCommand{\p}{p}
\DeclareBoldMathCommand{\u}{u}
\DeclareBoldMathCommand{\w}{w}
\DeclareBoldMathCommand{\x}{x}
\DeclareBoldMathCommand{\vzero}{0}
\let\top\intercal

\newcommand{\algp}{\widehat\p}
\newcommand{\textloss}{\textsc{loss}}
\newcommand{\squint}{\textsc{Squint}}
\newcommand{\metagrad}{\textsc{MetaGrad}}       
\newcommand{\clipmetagrad}{\textsc{MetaGrad+C}}
\newcommand{\restartgrad}{\textsc{MetaGrad+L}}
\newcommand{\newsquint}{\textsc{Squint+L}}
\newcommand{\clipsquint}{\textsc{Squint+C}}
\newcommand{\alg}{\textsc{Alg}}
\newcommand{\linregret}{\tilde{R}}            
\newcommand{\cliplinregret}{\bar{R}}         
\newcommand{\clipvar}{\bar{V}}               
\newcommand{\domain}{\mathcal{U}}
\newcommand{\simplex}{\triangle}             
\newcommand{\reals}{\mathbb{R}}
\DeclareMathOperator{\KL}{KL}
\DeclareMathOperator{\E}{\mathbb{E}}

\newcommand{\pred}{\widehat{\bm{u}}}
\let\footnotetitle\thanks
\DeclareRobustCommand{\VAN}[3]{#2} 

\makeatother
\begin{document}
\title[Lipschitz Adaptivity]{Lipschitz Adaptivity with Multiple Learning
Rates in Online Learning\protect\footnotetitle{Accepted for presentation at the Conference on Learning Theory (COLT) 2019}}
	
\coltauthor{\Name{Zakaria Mhammedi} \Email{zak.mhammedi@anu.edu.au} \\
  \addr School of Engineering and Computer Science \\
  The Australian National University and Data61 \\
  \Name{Wouter M.\ Koolen} \Email{wmkoolen@cwi.nl} \\
  \addr Centrum Wiskunde \& Informatica  \\ Amsterdam, the Netherlands\\
  \Name{Tim {van Erven}} \Email{tim@timvanerven.nl} \\
  \addr Statistics Department \\ Leiden University, the Netherlands
}

\maketitle

\begin{abstract}%
  We aim to design adaptive online learning algorithms that take
  advantage of any special structure that might be present in the
  learning task at hand, with as little manual tuning by the user as
  possible. A fundamental obstacle that comes up in the design of such
  adaptive algorithms is to calibrate a so-called step-size or learning
  rate hyperparameter depending on variance, gradient norms, etc. A recent
  technique promises to overcome this difficulty by maintaining multiple
  learning rates in parallel. This technique has been applied in the
  MetaGrad algorithm for online convex optimization and the Squint
  algorithm for prediction with expert advice. However, in both cases
  the user still has to provide in advance a Lipschitz hyperparameter that
  bounds the norm of the gradients. Although this hyperparameter is typically
  not available in advance, tuning it correctly is crucial: if it is set
  too small, the methods may fail completely; but if it is taken too
  large, performance deteriorates significantly. In the present work we
  remove this Lipschitz hyperparameter by designing new versions of
  MetaGrad and Squint that adapt to its optimal value automatically. We
  achieve this by dynamically updating the set of active learning rates.
  For MetaGrad, we further improve the computational efficiency of
  handling constraints on the domain of prediction, and we remove the
  need to specify the number of rounds in advance.
\end{abstract}

\section{Introduction}
\label{sec:intro}

We consider \emph{online convex optimization} (OCO) of a sequence of convex
functions $\ell_1,\ldots,\ell_T$ over a given bounded convex domain, which
become available one by one over the course of $T$ rounds
\citep{ShalevShwartz2011,HazanOCOBook2016}. Typically $\ell_t(\u) =
\textloss(\u,\x_t,y_t)$ represents the \emph{loss}
of predicting with parameters $\u$ on the $t$-th data point $(\x_t,y_t)$ in a machine
learning task. At the start of each round $t$, a learner has to predict
the best parameters $\pred_t$ for the function $\ell_t$ before finding out
what $\ell_t$ is, and the goal is to minimize the \emph{regret}, which is
the difference in the sum of function values between the learner's
predictions $\pred_1,\ldots,\pred_T$ and the best fixed oracle parameters $\u$
that could have been chosen if all the functions had been given in
advance. A special case of OCO is prediction with expert advice
\citep{cesa06}, where the functions $\ell_t(\u) = \tuple*{\u,
\vloss_t}$ are convex combinations of the losses $\vloss_t =
(\loss_{t,1},\ldots,\loss_{t,K})$ of $K$ expert predictors and the
domain is the probability simplex.

Central results in these settings show that it is possible to control the
regret with virtually no prior knowledge about the functions. For
instance, knowing only a $\norm{\cdot}_2$-upper-bound $G$ on the
gradients $\grad_t = \nabla \ell_t(\pred_t)$, the online gradient descent
(OGD) algorithm guarantees $O(G \sqrt{T})$ regret by tuning its learning
rate hyperparameter $\eta_t$ proportional to $1/(G\sqrt{t})$
\citep{Zinkevich2003}, and in the case of prediction with expert advice
the Hedge algorithm achieves regret $O(L\sqrt{T\ln K})$ knowing only an
upper-bound $L$ on the range $\max_k l_{t,k} - \min_k l_{t,k}$ of
the expert losses \citep{FreundSchapire1997}. Here $G$ is the
$\norm{\cdot}_2$-Lipschitz constant of the learning task\footnote{We slightly
abuse terminology here, because the standard definition of a Lipschitz
constant requires an upper-bound on the gradient norms for any
parameters $\u$, not just for $\u = \pred_t$, and may therefore be
larger.}, and $L/2$ is the $\norm{\cdot}_1$-Lipschitz constant over the
probability simplex.

The above guarantees are tight if we make no further assumptions about
the functions $(\ell_t)$ \citep{HazanOCOBook2016,CesaBianchiEtAl1997}, but
they can be significantly improved if the functions have additional
special structure that makes the learning task easier. The literature on
online learning explores multiple orthogonal dimensions in which tasks
may be significantly easier in practice (see `related work' below). Here,
we focus on the following refined data-dependent regret guarantees, which are known to exploit
multiple types of easiness at the same time:
\begin{eqnarray}
  \text{OCO:} & O\left(\sqrt{V_T^\u d \log T}\right)
  \text{ for all $\u$,}
  \quad &\text{with $V_T^\u = \sum_{t=1}^T \tuple*{\pred_t - \u,
  \grad_t}^2$,}\label{eqn:ourmetagradbound}\\
  \text{Experts:}& O\left(\sqrt{\E_{\rho(k)}[V_T^k]
  \KL(\rho\|\pi)}\right)
  \text{ for all $\rho$,}
  \quad &\text{with $V_T^k = \sum_{t=1}^T \tuple*{\pred_t - \e_k,
  \vloss_t}^2$,}\label{eqn:oursquintbound}
  \end{eqnarray}
where $d$ is the number of parameters and $\KL(\rho\|\pi) = \sum_{k=1}^K
\rho(k) \ln \rho(k)/\pi(k)$ is the Kullback-Leibler divergence from a fixed prior
distribution~$\pi$ over experts to any (data-dependent) comparator distribution $\rho$; for instance, $\rho$ is allowed here to be a
point-mass on the best expert $k^*$ in hindsight, in which case we would
have $\KL(\rho\|\pi) = -\ln \pi(k^*)$.

The OCO guarantee is achieved by the \metagrad{} algorithm
\citep{Erven2016}, and implies regret that grows at most logarithmically in
$T$ both in case the losses are curved (exp-concave, strongly convex)
and in the stochastic case whenever the losses are independent, identically
distributed samples with variance controlled by a Bernstein condition
\citep{koolen2016}. The guarantee for the expert case is
achieved by the \squint{} algorithm \citep{koolen2015,squintPAC}. It simultaneously exploits two types of structures: in many cases the $V_T^k$ term is much smaller than $L^2 T$ \citep{GaillardStoltzVanErven2014,koolen2016} and the so-called
\emph{quantile bound} $\KL(\rho\|\pi)$ is much smaller than the worst
case $\ln K$ when multiple experts make good predictions
\citep{ChaudhuriFreundHsu2009,ChernovVovk2010}. \squint{} and \metagrad{} are
both based on the same technique of tracking the empirical performance
of \emph{multiple learning rates} in parallel over quadratic
approximations of the original losses. A computational difference though is that \squint{}
is able to do this by a continuous integral that can be evaluated in
closed form, whereas \metagrad{} uses a discrete grid of learning rates.

Unfortunately, to achieve \eqref{eqn:ourmetagradbound} and
\eqref{eqn:oursquintbound}, both \metagrad{} and \squint{} need knowledge of
the Lipschitz constant ($G$ or $L$, respectively). Overestimating $G$ or
$L$ by a factor of $c > 1$ has the effect of reducing the effective
amount of available data by the same factor $c$, but underestimating the
Lipschitz constant is even worse since it can make the methods fail
completely. In fact, the ability to adapt to $G$ has been credited
\citep{WardWuBottou2018} as one of the main reasons for the practical
success of the AdaGrad algorithm
\citep{DuchiHazanSinger2011,McMahanStreeter2010}. Thus getting the
Lipschitz constant right makes the difference between having practical
algorithms and having promising theoretical results.

For OCO, an important first step towards combining Lipschitz adaptivity
to $G$ with regret bounds of the form \eqref{eqn:ourmetagradbound} was
taken by \citet{cutkosky2017}, who aimed for
\eqref{eqn:ourmetagradbound} but had to settle for a weaker result with
$G \sum_{t=1}^T \|\grad_t\|_2 \|\pred_t - \u\|_2^2$ instead of $V_T^\u$.
Although not sufficient to adapt to a Bernstein condition, they do
provide a series of stochastic examples where their bound already leads
to a fast $O(\ln^4 T)$ rates. For the expert setting,
\citet{Wintenberger2017} has made significant progress towards a version
of \eqref{eqn:oursquintbound} without the quantile bound improvement,
but he is left with having to specify an initial guess $L_\text{guess}$
for $L$ that enters as $O(\ln \ln (L/L_\text{guess}))$ in his bound, which may yet be arbitrarily large when the initial guess is on the
wrong scale.

\paragraph{Main Contributions.}

Our main contributions are that we complete the process began by
\citet{cutkosky2017} and \citet{Wintenberger2017} by showing that it is
indeed possible to achieve \eqref{eqn:ourmetagradbound} and
\eqref{eqn:oursquintbound} without prior knowledge of $G$ or $L$. In
fact, for the expert setting we are able to adapt to the tighter
quantity $B \geq \max_k \abs*{\tuple*{\pred_t - \e_k, \vloss_t}}$. We achieve these
results by dynamically updating the set of active learning rates in
\metagrad{} and \squint{} depending on the observed Lipschitz constants. In
both cases, we encounter a similar tuning issue as
\citet{Wintenberger2017}, but we avoid the need to specify any initial
guess using a new restarting scheme, which restarts the algorithm when
the observed Lipschitz constant increases too much. Interestingly, the
scheme and its analysis are different from the well-known doubling trick
\citep{cesa06}, and the regret bound is dominated by the regret incurred
over the last \emph{two} epochs instead of just the last epoch. 
Adding up the regret bounds over the last two epochs leads to at most an extra
$\sqrt{2}$ factor multiplying the final bound, and so this is the overhead we
incur
for Lipschitz adaptivity.
 In
addition to these main results, we remove the need to specify the number
of rounds $T$ in advance for \metagrad{} by adding learning rates as $T$
gets larger, and we improve the computational efficiency of how it
handles constraints on the domain of prediction: by a minor extension of
the black-box reduction for projections of \citet{cutkosky2018}, we
incur only the computational cost of projecting on the domain of
interest in \emph{Euclidean} distance. This should be contrasted with
the usual projections in time-varying Mahalanobis distance for
second-order methods like \metagrad{}.

\paragraph{Related Work.}
We build on several lines of work that achieve subsets of Lipschitz, variance and quantile adaptivity. Lipschitz adaptivity in OCO is achieved by OGD with learning rate
$\eta_t \propto 1/\sqrt{\sum_{s=1}^t \|\grad_s\|_2^2}$, which leads to
$O(\sqrt{\sum_{t=1}^T \|\grad_t\|_2^2}) = O(G \sqrt{T})$ regret. This
is the approach taken by AdaGrad (for each dimension separately)
\citep{DuchiHazanSinger2011,McMahanStreeter2010}. Lipschitz adaptive methods for prediction with
expert advice (sometimes called
\text{scale-free}) were obtained by
\citet{cbms07} and \citet{rooij14}.
These include a data-dependent variance term (though different from $V_T^k$ in \eqref{eqn:oursquintbound}), but no quantiles.

Dropping Lipschitz adaptivity, we find that bounds with $V_T^k$ from \eqref{eqn:oursquintbound} have previously been obtained
by \citet{GaillardStoltzVanErven2014} and \citet{Wintenberger2014Arxiv} without quantile bounds. Quantile adaptivity was achieved by \citet{ChaudhuriFreundHsu2009} and \citet{ChernovVovk2010} without variance adaptivity, and with a slightly weaker notion of variance by \citet{AdaNormalHedge}. In OCO, the analogue of quantile adaptivity is to adapt to the norm of $\u$, which has
been achieved in various different ways, see for instance
\citep{McMahanAbernethy2013,cutkosky2018}.

Several other important (and related) criteria of easiness are actively considered in the literature. These include curvature of the loss functions, where earlier results achieve fast rates assuming that the degree of
curvature is known \citep{HazanAgarwalKale2007}, measured online
\citep{BartlettHazanRakhlin2007,Do2009} or entirely unknown
\citep{Erven2016,cutkosky2018}. Fast rates are also possible for
slowly-varying linear functions and, more generally, optimistically
predictable gradient sequences
\citep{hazan2010extracting,GradualVariationInCosts2012,RakhlinSridharan2013}.

We view our results as a step towards developing algorithms that
automatically adapt to multiple relevant measures of difficulty at the
same time. It is not a given that such combinations are always possible.
For example, \citet{CutkoskyBoahen2017Impossible} show that Lipschitz
adaptivity and adapting to the comparator complexity in OCO, although
both achievable independently, cannot both be realized at the same time
(at least not without further assumptions). A general framework to study
which notions of task difficulty do combine into achievable bounds is
provided by \citet{FosterRakhlinSridharan2015}.
\citet{FosterRakhlinSridharan2017} characterize the achievability of
general data-dependent regret bounds for domains that are balls in
general Banach spaces. 

\paragraph{Outline.}

We add Lipschitz adaptivity to \squint{} for the expert setting in
Section~\ref{Squint2}. Then, in Section~\ref{MetaC}, we do the same
for \metagrad{} in the OCO setting. The developments are analogous at a
high level but differ in the details for computational reasons. We
highlight the differences along the way. Section~\ref{MetaC} further
describes how to avoid specifying $T$ in advance for \metagrad{}. Then, in
Section~\ref{four}, we add efficient projections for \metagrad{}, and
finally Section~\ref{sec:conclusion} concludes with a discussion of
directions for future work.

\paragraph{Problem Setting and Notation.}

In OCO, a learner repeatedly chooses actions $\pred_t$ from a closed convex
set $\domain \subseteq \reals^d$ during rounds $t=1,\ldots,T$, and
suffers losses~$\ell_t(\pred_t)$, where $\ell_t: \domain \to \reals$ is a convex
function. The learner's goal is to achieve small regret
%
  $R_T^\u = \sum_{t=1}^T \ell_t(\pred_t) - \sum_{t=1}^T \ell_t(\u)$
%
with respect to any comparator action $\u \in \domain$, which measures
the difference between the cumulative loss of the learner and the
cumulative loss they could have achieved by playing the oracle
action~$\u$ from the start. A special case of OCO is prediction with
expert advice, where $\ell_t(\u) = \tuple*{\u, \vloss_t}$ for $\vloss_t \in
\reals^K$ and the domain $\domain$ is the probability simplex
$\simplex_K = \{(u_1,\ldots,u_K) : u_i \geq 0, \sum_i u_i = 1\}$.
In this context we will further write $\p$ instead of $\u$ for the
parameters to emphasize that they represent a probability distribution.
We further define $[K] = \{1,\ldots,K\}$. 

\section{An Adaptive Second-order Quantile Method for Experts}
\label{Squint2}

In this section, we present an extension of the \squint{} algorithm that adapts automatically to the loss range in the setting of prediction with expert advice. 

Throughout this section, we denote the \emph{instantaneous regret} of expert $k\in[K]$ in round $t$ by $r^k_t \coloneqq \inner{\what{\bm{p}}_t- \bm{e}_k}{\vloss_t}$, where $\bh{p}_t \in \triangle_K$ is the weight vector played by the algorithm and $\vloss_t\in \reals^K$ is the observed loss vector. The cumulative regret with respect to expert $k$ is given by $\tmp{R}^k_t\coloneqq \sum_{s=1}^t r^k_s$. 
The cumulative `variance' with respect to expert $k$ is measured by
$\tmp{V}^k_t \coloneqq \sum_{s=1}^t v^k_s$ for $v^k_t \coloneqq
(r^k_t)^2$. In the next subsection, we review the \squint{} algorithm.

\subsection{The \squint{} Algorithm}
\label{AdaptiveSquint}

We first describe the original 
\squint{} algorithm as introduced by \cite{koolen2015}. Let $\pi$
and $\gamma$ be prior distributions with supports on $k \in [K]$ and
$\eta \in \left]0,
1/2\right]$, respectively. After $t$ rounds,
\squint{} outputs predictions
\begin{gather}
\label{Squintforcaster}
  \algp_{t+1} \propto \underset{\pi(k)\gamma(\eta)}{\mathbb{E}}\left[ \eta e^{-  \sum_{s=1}^t f_s(k,\eta)} \bm{e}_k \right],
\shortintertext{where $f_t(k,\eta)$ are quadratic \emph{surrogate losses} defined by}
\nonumber
f_t(k,\eta) \coloneqq - \eta \inner{\bh{p}_t-\bm{e}_k}{\vloss_t} + \eta^2 \inner{\bh{p}_t-\bm{e}_k}{\vloss_t}^2.
\end{gather} 
\cite{koolen2015} propose to use the \emph{improper prior} $\gamma(\eta)
= 1/\eta$ which does not integrate to a finite value over its
domain, but because of the weighting by $\eta$ in
\eqref{Squintforcaster} the predictions $\algp_{t+1}$ are still
well-defined. The benefit of the improper prior is that it allows
calculating $\algp_{t+1}$ in closed form \citep{koolen2015}. It is also
the natural candidate for Lipschitz adaptivity, as it is scale-invariant: the density of an interval only depends on the ratio of its endpoints, not on their location. For any distribution $\rho \in \simplex_K$, \squint{} achieves the following bound:
\begin{align}
\tmp{R}^{\rho}_T = O\left(\sqrt{\tmp{V}^{\rho}_T\left( \KL(\rho || \pi ) +
\ln \ln T\right)}\right), \nonumber
\end{align} 
where $R_T^{\rho} = \mathbb{E}_{\rho(k)}\left[R_T^{k} \right]$ and
$V_T^{\rho} = \mathbb{E}_{\rho(k)}\left[V_T^{k} \right]$. This version
of \squint{} assumes the loss range $\max_k \loss_{t,k} - \min_k \loss_{t,k}$
is at most $1$, and can fail otherwise. In the next subsection, we
present an extension of \squint{} which does not need to know the
Lipschitz constant.

\subsection{Lipschitz Adaptive \squint{}}

\let\scale\bar
We first design a version of \squint{}, called \clipsquint{}, that still requires an
initial estimate $B$ of the Lipschitz constant. We then present
\newsquint{} which tunes this parameter online. For now, we consider a fixed $B>0$. In addition to this, the algorithm takes a prior distribution $\pi \in \triangle_K$. 
We denote the observed Lipschitz constant in round $t$ at the algorithm's prediction $\algp_t$ by
$
b_t
\df
\max_k \abs{r_t^k} = \max_k |\tuple{\algp_t - \e_k, \vloss_t}|$, and denote its running maximum by $B_t \df B \lub
\max_{s \le t} b_s$, with the convention that $B_0=B$. We will also require a \emph{clipped} version of the loss vector
$\scale{\vloss}_t = \vloss_t  \cdot B_{t-1}/B_t$, and denote by
$\scale{r}_t^k = \tuple{\algp_t - \e_k, \scale \vloss_t}$ the
\emph{clipped instantaneous regret}; we will use that $\abs{\scale r_t^k} \le
B_{t-1}$. Following \citet{cutkosky2019artificial}, it suffices to control the regret for the clipped loss,
because the cumulative difference is of the order of one round (\emph{i.e.}\ a negligible lower-order constant):
\begin{equation}\label{eq:ashok}
  R_T^k
  -
  \scale R_T^k
  ~\df~
  \sum_{t=1}^T \del*{r_t^k - \scale r_t^k}
  ~=~
  \sum_{t=1}^T \del*{B_t - B_{t-1}} \frac{r_t^k}{B_t}
  ~\le~
  B_T - B_0
  .
\end{equation}
This means we can focus on the regret for $\scale \vloss_t$, for which the range bound $\abs{\scale r_t^k} \le B_{t-1}$ is available \emph{ahead} of each round $t$. To motivate \clipsquint{}, we define the potential function after $T$ rounds by
\begin{equation}\label{eq:sq.pot}
  \Phi_T
  \df
  \sum_k \pi_k \int_0^\frac{1}{2 B_{T-1}} \frac{e^{\eta \scale{R}_T^k - \eta^2 \scale{V}_T^k} -1}{\eta} \dif \eta
  \quad
  \text{where}
  \quad
  \scale R_T^k \df \sum_{t=1}^T \scale r_t^k
  ~~
  \text{and}
  ~~
  \scale V_T^k \df \sum_{t=1}^T (\scale r_t^k)^2
  .
\end{equation}
We also define $\Phi_0 = 0$ (due to the integrand being zero), even though it involves the meaningless $B_{-1}$ in the upper limit. The algorithm is now derived from the desire of keeping this potential under control. As we will see in the analysis, this requirement uniquely forces the choice of weights
\begin{equation}\label{eq:sq.weights}
  \widehat p_{T+1}^k
  ~\propto~
  \pi_k \int_0^\frac{1}{2 B_T} e^{\eta \scale{R}_T^k - \eta^2 \scale{V}_T^k} \dif \eta
  .
\end{equation}
The predictions $\algp_{t+1}$ take the same functional form as the original \squint{}, and can hence be evaluated in closed form (\emph{i.e.}\ in terms of the Gaussian CDF).
 The regret analysis consists of two parts. First, we show that the algorithm keeps the potential small:
\begin{lemma}\label{lem:pot.is.small}
Given parameter $B > 0$, \clipsquint{} ensures $\Phi_T \le \ln
\frac{B_{T-1}}{B}$.
\end{lemma}
The next step of the argument is to show that a small potential $\Phi_T$
is useful. The argument here follows from \citep{koolen2015},
specifically the version by \cite{squintPAC}. We have:
\begin{lemma}\label{lem:small.is.good}
  For any comparator
  distribution $\rho \in \triangle_K$ the regret of \clipsquint{} is at most
  \begin{gather}\nonumber 
    \scale R_T^\rho
     ~\le~
    \sqrt{2     \scale V_T^\rho} \del*{
      1+
      \sqrt{2 C_T^{\rho}}
    }
    +
    5 B_{T-1}
    \del*{C_T^{\rho}+ \ln 2}, 
    \quad   \text{where}  \\
    C^{\rho}_T ~\df~
    \KL \delcc*{\rho}{\pi}
    + \ln \del*{
      \Phi_T
      + \frac{1}{2}
      + \ln \left(2+  \sum_{t=1}^{T-1} \frac{b_t}{B_t} \right)\nonumber 
    }.
  \end{gather}
\end{lemma}
Keeping only the dominant terms, this reads
\(
  \scale R_T^\rho
  =
  O\del*{
    \sqrt{\scale V_T^\rho \del*{\KL \delcc*{\rho}{\pi} +
        \ln \del*{\Phi_T + \ln T}}}
  }
\).
Combining with \eqref{eq:ashok}, and Lemmas~\ref{lem:pot.is.small}
and~\ref{lem:small.is.good}, we obtain a bound of the form
\begin{align}
  R_T^\rho
  ~=~
  O \del*{
    \sqrt{V_T^\rho \del*{\KL \delcc*{\rho}{\pi} +
   \ln    \ln \frac{TB_{T-1}}{B}}}
  +
  5 B_T \del*{\KL \delcc*{\rho}{\pi} +
  \ln   \ln \frac{TB_{T-1}}{B} }
}
  . \label{eq:regret}
\end{align}
However, there does not seem to be any safe a-priori way to tune
$B=B_0$. If we set it too small, the factor $\ln \ln (B_{T-1}/B)$
explodes. If we set it too large, with $B$ much larger
than the effective range of the data, then $B_T = B$ and the
term outside the square-root on the RHS of \eqref{eq:regret} blows
up. It does not appear possible to bypass this tuning dilemma directly
within the current construction. Instead, we solve this problem using a
new type of restarts that are different from the well-known
doubling trick. For this, we present Algorithm~\ref{bb1alg}, which applies to both \clipsquint{} and \clipmetagrad{} (presented in the next section). It monitors a condition on the sequences $(b_t)$ and $(B_t)$ to trigger restarts.
\begin{algorithm}[tbp]
\caption{Restarts to make {\clipsquint} or {\clipmetagrad} scale-free.}
\label{bb1alg}
\begin{algorithmic}[1]
\REQUIRE {\alg} is either {\clipsquint} or {\clipmetagrad}, taking as
input parameter an initial scale $B$;
\STATE Play $\vzero$ for OCO or $\pi$
for experts until the first time $t=\tau_1$ that $b_t \neq 0$; 
\STATE \label{line:runmetagrad} Run {\alg} with input $B = B_{\tau_1}$ until the first time $t=\tau_2$ that $\displaystyle \frac{B_t}{B_{\tau_1}} > \sum_{s=1}^t \frac{b_s}{B_s}$;\\
\STATE Set $\tau_1 = \tau_2$ and goto line \ref{line:runmetagrad};
\end{algorithmic}
\end{algorithm}
\begin{theorem}
  \label{blackboxreduction0}
  Let \newsquint{} be the result of applying
Algorithm~\ref{bb1alg} with \clipsquint{} as \textsc{Alg}. \newsquint{} guarantees, for any comparator $\rho\in \triangle_K$,
\begin{align}
  R_T^\rho
    ~\le~
    2\sqrt{      V_T^\rho} \del*{
      1+
      \sqrt{2 \Gamma_T^{\rho}}
    }
    +
    10 B_{T}
    \del*{
      \Gamma_T^{\rho}
      + \ln 2} + 4 B_T, \nonumber
\end{align}
where $  \Gamma_T^{\rho} \df\KL \delcc*{\rho}{\pi}+ \ln \del*{\ln
\left(\sum_{t=1}^{T-1} b_{t}/B_t\right) + \ln
\left(2+\sum_{t=1}^{T-1} b_{t}/B_t\right)}+ 1/2$.
\end{theorem}
Note that $\Gamma_T^{\rho}$ in Theorem \ref{blackboxreduction0} is equal
to $\KL \delcc*{\rho}{\pi} + O\del*{\ln \ln T}$. Importantly, this
theorem and Algorithm~\ref{bb1alg} do not depend on any initial guess
$B$ anymore. Instead, Algorithm~\ref{bb1alg} plays the starting
parameters until the first time a non-zero loss is observed, and then
monitors a data-dependent criterion that measures whether the loss range
has increased by more than a factor that is roughly~$t$, to decide when
to trigger a restart. For most types of data, such large increases in
the loss range should be rare after a few start-up rounds, so restarts
should quickly stop occurring.

\section{An Adaptive Method for Online Convex Optimization}
\label{MetaC}
We now present an extension of the \metagrad{} algorithm which adapts automatically to the gradient norm in online convex optimization --- we call this Lipschitz adaptive version \restartgrad{}.
Recall that in the OCO setting, at each round $t$, the learner predicts
a vector $\what{\bm{u}}_t$ in a closed convex set $\mathcal{U} \subset
\mathbb{R}^d$, then suffers loss $\ell_t(\what{\bm{u}}_t)$, where $\ell_t :
\mathcal{U}\rightarrow \reals$ is a convex function. The goal of
the learner is to minimize the regret $\tmp{R}^{\bm{u}}_T \coloneqq
\sum_{t=1}^T \ell_t(\what{\bm{u}}_t) - \sum_{t=1}^T \ell_t(\bm{u})$ with
respect to the single best action $\bm{u}\in \mathcal{U}$ in hindsight.
In this case, convexity of the losses implies that $\ell_t(\what{\bm{u}}_t)
- \ell_t(\bm{u}) \leq \inner{\what{\bm{u}}_t - \bm{u}}{\bm{g}_t}$, where
$\bm{g}_t \coloneqq \nabla \ell_t(\what{\bm{u}}_t)$, and so it suffices to
control the \emph{pseudo-regret} $\tilde{R}^{\bm{u}}_T \df
\sum_{t=1}^T \inner{\what{\bm{u}}_t - \bm{u}}{\bm{g}_t}$. We will assume that the set $\mathcal{U}$ is bounded, and denote its diameter by
\begin{align} \label{rad}D \coloneqq \sup_{\bm{u}, \bm{v}\in \mathcal{U}} \norm{\bm{u} - \bm{v}}_2.\end{align}
Without loss of generality, we will also assume that the set $\mathcal{U}$ is centered at $\bm{0}$.
The proofs for this section are deferred to Appendix \ref{MetaGrad2proofs}. We now review the \metagrad{} algorithm.

 \subsection{The \metagrad{} Algorithm}
 The \metagrad{} algorithm runs several sub-algorithms at each round: namely, a set of slave algorithms, which learn the best action in $\mathcal{U}$ given a learning rate $\eta$ in some pre-defined grid $\mathcal{G}$, and a master algorithm, which learns the best learning rate. Through this, the \metagrad{} algorithm controls the sum of \emph{surrogate losses} $\sum_{t=1}^T f_t(\bm{u},\eta)$ over all $\eta \in \mathcal{G}$ and $\bm{u}\in \mathcal{U}$ simultaneously, where
 \begin{align}
\label{surrmeta}
f_t(\bm{u},\eta) \coloneqq - \eta \inner{\bh{u}_t -   \bm{u}}{\bm{g}_t}  +  \eta^2 \inner{\bh{u}_t -   \bm{u}}{\bm{g}_t}^2,
\end{align}
and $\bh{u}_t$ is the master's prediction at round $t\in  [T]$. Each slave algorithm takes as input a learning rate from a finite grid $\mathcal{G}$ (with $\ceil{1/2  \log_2 T}$ points) in the form of a geometric progression and within the interval $[1/(5DG\sqrt{T}), 1/(5DG)]$, where $G$ is an upper-bound on the norms of the gradients. In this case, $G$ must be known in advance to construct the grid; in the proof of \metagrad's regret bound, it is crucial for the learning rates to be in the right interval in order to invoke a certain Gaussian exp-concavity result due to \citet{Erven2016} for the surrogate losses in \eqref{surrmeta}. In what follows, we let $\mathbf{S}_t \coloneqq \sum_{s=1}^t \bm{g}_s\bm{g}_s^{\T}$, for $ t \geq 0$.
\paragraph{Slaves' Predictions.}
Each slave $\eta \in \mathcal G$ starts with $\bh{u}_1^\eta = \bm{0} \in \mathcal U$, and at the end of round $t \ge 1$, it receives the master's prediction $\bh{u}_t$ and updates its own prediction in two steps:
\begin{gather}
   {\bm{u}}^{\eta}_{{t+1}} \coloneqq \bh{u}^{\eta}_t - \eta \mathbf{\Sigma}^{\eta}_{{t+1}} \bm{g}_t  \left(1 + 2 \eta \left( \bh{u}^{\eta}_t -\bh{u}_t \right)^{\T}\bm{g}_t\right) , \text{ where }\ \mathbf{\Sigma}^{\eta}_{{t+1}} \coloneqq \left( \tfrac{\mathbf{I}}{D^2} +2\eta^2\mathbf{S}_t \right)^{-1},
                            \label{quadprog0}\\
                            \nonumber
   \text{ and } \ \ \bh{u}^{\eta}_{{t+1}} = \argmin_{\bm{u}\in \mathcal{U}} \left(\bm{u}_{{t+1}}^{\eta}- \bm{u} \right)^{\T}\left( \mathbf{\Sigma}^{\eta}_{{t+1}}\right)^{-1} \left(\bm{u}_{{t+1}}^{\eta}- \bm{u} \right).
\end{gather}
\paragraph{Master's Predictions.}
After receiving the slaves' predictions, $\left(\bh{u}^{\eta}_t\right)_{\eta \in \mathcal{G}}$, at round $t\ge 1$, the master algorithm aggregates them and outputs $\bh{u}_t\in \mathcal{U}$ according to: 
\begin{align}
\bh{u}_{t} \coloneqq \frac{\sum_{\eta\in \mathcal{G}}  \eta w^{\eta}_t   \bh{u}^{\eta}_{t} }{\sum_{\eta \in \mathcal{G}}\eta w^{\eta}_t };\quad w^{\eta}_t \coloneqq e^{- \sum_{s=1}^{t-1} f_s(\bh{u}^{\eta}_s,\eta)}. \nonumber
\end{align}
\citet{Erven2016} showed that \metagrad{} has regret bounded by \eqref{eqn:ourmetagradbound}. In the next subsection, we present an extension of \metagrad{} which does not require knowledge of either the horizon $T$ or the Lipschitz constant (\emph{i.e.} a bound on the norms of the gradients).

\subsection{Lipschitz Adaptive \metagrad{}}
Similar to the \squint{} case, we first design a version of \metagrad{}, called \clipmetagrad{}, which still requires an input $B > 0$ (in this case, $B/D$ is the initial estimate of the Lipschitz bound). We then present \restartgrad{} which sets this parameter online. For now, we consider a fixed $B>0$. We define $
b_t
\df D \norm{\nabla \ell_t(\pred_t)}_2
= D \norm{\grad_t}_2$, for $t\geq 1$, and $b_0 \coloneqq B$. We denote the running maximum of $(b_t)$ by $B_t \df \max_{0\le s \le t} b_s$. We will also require a \emph{clipped} version of the gradient vector $\scale{\grad}_t = \grad_t \cdot  B_{t-1}/B_t$, and denote by $\scale{r}_t^\u = \tuple{\pred_t - \u, \scale\grad_t}$ the \emph{clipped instantaneous pseudo-regret} with respect to $\u \in \mathcal{U}$. In addition, it will be useful to define
\begin{align}
\bar{f}_t(\bm{u},\eta)\coloneqq - \eta \bar{r}^{\bm{u}}_t + \left(\eta \bar{r}^{\bm{u}}_t\right)^2 \quad   \text{and} \quad  \scale{\mathbf{S}}_t \coloneqq \sum_{s=1}^t\scale{\grad}_s \scale \grad_s^{\top}. \label{clippedstuff}
\end{align}
Recall that in the original \metagrad{}, the horizon $T$ and the Lipschitz constant $G$ were required to construct the grid of learning rates. We circumvent this by defining an infinite grid $\mathcal{G}$ in which, at any given round $t\geq1$, only a finite number of (active) slaves  --- up to $\log_2 t$ many --- output a non-zero prediction. Each slave $\eta$ in this grid receives a prior weight $\pi(\eta) \in[0,1]$, where $\sum_{\eta\in \mathcal{G}} \pi(\eta) =1$. Given input $B>0$ to \clipmetagrad, the grid $\mathcal{G}$ and the prior $\pi$ are defined by
\begin{align}
\mathcal{G}  \coloneqq \left\{ \eta_i \coloneqq   \frac{1}{5 B2^{i}}: i \in \mathbb{N} \cup \{0\}  \right\} \label{Ggrid} ;\quad 
\pi(\eta_i) \coloneqq  \frac{1}{(i+1)(i+2)},  \ i\in \mathbb{N}\cup \{0\}.
\end{align}
The subset of active slaves $\mathcal{A}_t$ at a round $t\geq 1$ is given by 
\begin{align}
\mathcal{A}_t \coloneqq \left\{ \eta \in \mathcal{G}\cap \left[0,\tfrac{1}{5B_{t-1}}\right] : s_\eta <   t   \right\}, \text{ with }
  s_\eta
  \df
  \min   \left\{t \ge 0 :
  \frac{1}{\eta}    \leq D \sum_{s=1}^t \norm{\bar{\bm{g}}_s}_2 + B_t  \label{threshold}
  \right\}.
\end{align}
We note that restricting the slaves (or learning rates) to the set $\mathcal{G}_t \coloneqq \mathcal{G} \cap [0,1/(5B_{t-1})]$ is similar in principle to clipping the upper integral range in the \clipsquint{} case.

\paragraph{Slaves' Predictions.} A slave $\eta \in \mathcal{G} \cap [0,1/(5B_{t-1})]$ issues predictions $\bh{u}_t^\eta = \bm{0}$ in all rounds $t\leq s_\eta+1$. From then on (\emph{i.e.} at the end of round $t\geq s_{\eta}+1$), it receives the master's prediction $\bh{u}_t$ as input and updates its own prediction in two steps:
\begin{gather}
  \bm{u}^{\eta}_{{t+1}} \coloneqq \bh{u}^{\eta}_t - \eta \mathbf{\Sigma}^{\eta}_{{t+1}} \bar{\bm{g}}_t  \left(1 + 2 \eta \left( \bh{u}^{\eta}_t -\bh{u}_t \right)^{\T}\bar{\bm{g}}_t\right),
                          \text{ where }\
  \mathbf{\Sigma}^{\eta}_{{t+1}} \coloneqq \left( \tfrac{\mathbf{I}}{D^2} +2\eta^2\left(\bar{\mathbf{S}}_t -\bar{\mathbf{S}}_{s_{\eta}}\right) \right)^{-1},
                          \nonumber
  \\
 \text{ and }\ \ \bh{u}^{\eta}_{{t+1}} = \argmin_{\bm{u}\in \mathcal{U}} \left(\bm{u}_{{t+1}}^{\eta}- \bm{u} \right)^{\T}\left( \mathbf{\Sigma}^{\eta}_{{t+1}}\right)^{-1} \left(\bm{u}_{{t+1}}^{\eta}- \bm{u} \right).
                         \nonumber
\end{gather}

 \paragraph{Master's Predictions.} At each round $t\geq1$, the  master algorithm receives the slaves' predictions $(\bh{u}_t^{\eta})_{t\in \mathcal{A}_{t}}$ and outputs
 \begin{align}
\label{newmaster}
\bh{u}_t = \frac{\sum_{\eta \in \mathcal{A}_{t}}\eta   w^{\eta}_t \bh{u}_t^{\eta}}{\sum_{\eta \in \mathcal{A}_{t}} \eta  w^{\eta}_t  }, \quad \text{where} \ \   w^{\eta}_t \coloneqq \pi(\eta)  e^{- \sum_{s=s_{\eta}+1}^{t-1} \bar{f}_s(\bh{u}^{\eta}_s,\eta)}.
\end{align}
\begin{remark}[Number of Active Slaves]  \label{numslaves} At any round $t\geq 1$, the number of active slaves is at most $\ceil{\log_2 t}$. In fact, if $\eta \in \mathcal{A}_t$, then by definition $\eta \geq 1/(D\sum_{s=1}^{s_{\eta}}\norm{\bm{g}_s}_2 + B_{s_{\eta}}) \geq 1/(t B_{t-1})$ (since $s_{\eta}\leq t-1$), and thus $\mathcal{A}_t \subset [1/(tB_{t-1}), 1/(5B_{t-1})]$.
Since $\mathcal{A}_t$ is a grid in the form of a geometric progression with common ratio $2$, there are at most $\ceil{\log_2 t}$ slaves in $\mathcal{A}_t$.
\end{remark}
To motivate \clipmetagrad{}, we define the potential function after $t\geq0$ rounds by\begin{align}
\label{masterpot}
 \Phi_t \coloneqq  \pi(\mathcal{G}_t\setminus \mathcal{A}_t) +  \sum_{\eta\in \mathcal{A}_t}  \pi(\eta)  e^{-\sum_{s=s_{\eta}+1}^t \bar{f}_s(\bh{u}^{\eta}_s,\eta)}, \quad \text{where } \mathcal{G}_t \coloneqq \mathcal{G}\cap \left[0,\tfrac{1}{5B_{t-1}}\right].\end{align}
Let $\bm{u}\in \mathcal{U}$. Recall that the pseudo-regret is defined by $\tilde{R}^{\bm{u}}_T  \coloneqq \sum^T_{t=1} \inner{\bh{u}_t - \bm{u}}{{\bm{g}}_t}$. We now defined its \emph{clipped} version by $\cliplinregret^{\bm{u}}_T  \coloneqq \sum^T_{t=1} \inner{\bh{u}_t - \bm{u}}{\bar{\bm{g}}_t}$. For $r^{\u}_t \coloneqq \inner{\pred_t - \u}{\grad_t}$, we have, similarily to \eqref{eq:ashok}, 
\begin{equation}\label{clippedrel}
  \tilde{R}_T^\u
  -
  \scale R_T^\u
 =
  \sum_{t=1}^T \del*{r_t^\u - \scale r_t^\u}
  ~=~
  \sum_{t=1}^T \del*{B_t - B_{t-1}} \frac{r_t^\u}{B_t}
  ~\le~
  B_T - B_0
  ,
\end{equation}
where the last inequality follows from the Cauchy-Schwarz inequality and the fact that $\mathcal{U}$ has diameter $D$, which together imply that $|r^{\bm{u}}_t| \leq B_t$. Using the inequality $e^{x-x^2}-1\leq x$, which holds for all $x\geq -1/2$, one can shown that the potential is a decreasing function of the number of rounds:
 \begin{lemma}
\label{lemmameta}
\clipmetagrad{} guarantees that $\Phi_T \leq \dots \leq  \Phi_0 = 1$, for all $T \in \mathbb{N}$.
\end{lemma}
We now give an upper-bound on $\cliplinregret^{\u}_T$ in terms of the clipped `variance' $\clipvar^{\u}_T \coloneqq \sum_{t=1}^T (\bar{r}^{\u}_t)^2$;
\begin{theorem}
\label{naivemeta}
Given input $B>0$, the clipped pseudo-regret for {\clipmetagrad} is bounded by
\begin{equation}
\cliplinregret_T^\u\leq
3\sqrt{\clipvar_T^\u C_T} + 15 B_T  C_T,
\quad \text{for any $\u \in \domain$,}\nonumber
\end{equation}
where $C_T \coloneqq  d\ln\left(1 + \frac{2  \sum_{t=0}^{T-1}
b_t^2 }{25 d B^2_{T-1}}\right) +  2 \ln \left( \log^+_2
\frac{\sqrt{\sum_{t=1}^Tb^2_t }}{B} +3 \right) + 2$ and $\log_2^+ = 0 \vee \log_2 $.
\end{theorem}
\begin{remark} 
\label{truebound}
For $\u \in \mathcal{U}$, we can relate the clipped pseudo-regret to the ordinary regret 
via $R_T^\u \leq \linregret_T^\u \leq \cliplinregret_T^\u + B_T$ (see
\eqref{clippedrel}) and on the
right-hand side we can also use that $\clipvar_T^\u \leq V_T^\u$.
\end{remark} 
An important aspect to note from Theorem \ref{naivemeta} is that the ratio $\sqrt{\sum_{t=1}^Tb^2_t}/B$, could in principle be arbitrarily large if the input $B$ is too small compared to the actual norms of the gradients (for \squint{} it was the ratio $B_{T-1}/B$ which was problematic). To resolve this issue, we use the same restart approach as in the \squint{} case:
\begin{theorem}
\label{blackboxreduction1}
Let {\restartgrad} be the result of applying Algorithm~\ref{bb1alg} to
{\clipmetagrad}. Then the actual and linearised regrets for {\restartgrad} are both bounded by
\begin{align}
  R^{\bm{u}}_t
  ~\le~
   \linregret_T^\u 
  ~\leq~
  3\sqrt{V_T^\u \Gamma_T} + 15 B_T \Gamma_T + 4 B_T
    \quad \text{for all $\u \in \domain$,}\nonumber
\end{align}
where $\Gamma_T \coloneqq  2 d\ln\left(1 + \frac{2}{25d}  \sum_{t=1}^{T} \frac{b_t^2}{B^2_{t}}\right) +  4 \ln \left( \log^+_2
\sqrt{\sum_{t=1}^T (\sum_{s=1}^t \frac{b_s}{B_s})^2} +3 \right)+ 4 = O(d
\ln T)$.
\end{theorem}
Theorem \ref{blackboxreduction1} replaces the ratio
$\sqrt{\sum_{t=1}^Tb^2_t} /B$ appearing in the (clipped) pseudo-regret bound of \clipmetagrad{} by $\sigma_T
\coloneqq \sqrt{\sum_{t=1}^T (\sum_{s=1}^t b_s/B_s)^2}$. The latter is independent of the input $B$ and is
always smaller than $T^{3/2}$; this is perfectly affordable since $\sigma_T$
appears inside a $\ln \ln$. Our reason for including the linearised
regret $\linregret_T^\u$ in Theorem~\ref{blackboxreduction1} is that a bound on it in terms of $V_T^\u$ is the precondition for fast rate results in individual-sequence settings based on curvature \citep{Erven2016} and in statistical settings under certain (Bernstein type) conditions \citep{koolen2016}.

 \section{Efficient Implementation Through a Reduction to the Ball}
 \label{four}
 Using \metagrad{} (\textsc{+C} or \textsc{+L}), the computation of
 each slave prediction $\bh{u}^{\eta}_t$ requires a projection onto an
 arbitrary convex set $\mathcal{U}$ in \emph{Mahalanobis distance}. Numerically, this typically
 requires $O(d^p)$ floating point operations (flops), for some $p \in
 \mathbb{N}$ which depends on the geometry of the set $\mathcal{U}$.
 Since $p$ can be large in many applications,
 evaluating $\bh{u}^{\eta}_{t}$ for each slave $\eta$ can become computationally prohibitive, especially when the number of slaves grows with $T$; for the \metagrad{} versions discussed in this paper, there can be up to $\ceil{\log_2 T}$ slaves at round $T\geq1$ (see Remark \ref{numslaves}).

 The goal of this section is to streamline these computations, which we will do in two steps. In Section~\ref{ballefficient}, we will describe an efficient implementation of \metagrad{} on the ball. The main idea here is that the Mahalanobis projections onto the ball, which are performed by the slaves, can reuse a common matrix decomposition. In Section~\ref{bbred.gnrlzd}, we will then obtain an algorithm for any bounded convex set $\mathcal U$ by applying the black-box reduction of \cite{cutkosky2018} to \metagrad{} on the ball enclosing $\mathcal U$. We show (Theorem \ref{reductionbound}) that the reduction also transports variance bounds.
 The techniques discussed here also apply to the versions of \metagrad{} presented in the previous section. However, to simplify the presentation, we will only
focus on the original \metagrad{}. The proofs for this section are
deferred to Appendix \ref{fourproof}.
 \subsection{Efficient Implementation of \metagrad{} on the Ball}
 \label{ballefficient}
 Suppose that $\mathcal{U}$ is the ball of diameter $D$:
 $\mathcal{U}=\mathcal{B}_{D} \coloneqq \left\{\bm{u} \in \mathbb{R}^d
 \colon \norm{\bm{u}}_2 \leq D/2 \right\}$. To compute the slave's
 prediction $\bh{u}^{\eta}_{t+1}$, the following quadratic program needs
 to be solved for each $\eta$:
\begin{align}
\bh{u}^{\eta}_{t+1} = \argmin_{\bm{u}\in \mathcal{U}} \left(\bm{u}_{t+1}^{\eta}- \bm{u} \right)^{\T}\left( \mathbf{\Sigma}^{\eta}_{t+1}\right)^{-1} \left(\bm{u}_{t+1}^{\eta}- \bm{u} \right), \label{quadprog2}
\end{align}   
where $\bm{u}^{\eta}_{t+1}$ (the unprojected prediction) and
$\mathbf{\Sigma}^{\eta}_{t+1} = (
\mathbf{I}/D^2 + 2\eta^2 \mathbf{S}_t)^{-1}$ (the co-variance matrix) are defined in
\eqref{quadprog0}. Since $\mathcal{U}$ is a ball and
$\mathbf{\Sigma}^{\eta}_{t+1}$ is symmetric positive-definite,
\eqref{quadprog2} can be solved in $O(d^3)$ by performing a singular
value decomposition of $\mathbf{\Sigma}^{\eta}_{t+1}$. Instead of doing
this singular value decomposition separately for each $\eta$, we can be
a little more efficient by doing a singular value
decomposition of $\mathbf{S}_t$ once and then using the following lemma:
\begin{lemma}
\label{redquad}
Let
$\mathbf{\Lambda}_t \coloneqq \op{diag}((\lambda^i_t)_{i\in[d]})$ and
$\mathbf{Q}_t$ be the matrices of eigenvalues and eigenvectors of
$\mathbf{S}_t$, respectively, such that $\mathbf{Q}_t {\mathbf{S}}_t
\mathbf{Q}^{\T}_t = \mathbf{\Lambda}_t$ and $\mathbf{Q}_t
\mathbf{Q}_t^\top = \mathbf{I}$.\footnote{The existence of such a
$\mathbf{Q}_t$ and $\mathbf{\Lambda}_t$ is guaranteed due to
${\mathbf{S}}_t$ being symmetric positive-definite.}
Then the solution of \eqref{quadprog2} is
\[
  \bh{u}^{\eta}_{t+1} =
  \begin{cases}
    \bm{u}^{\eta}_{t+1},
      & \text{if $\bm{u}^{\eta}_{t+1} \in \mathcal{U}$,}\\
    \mathbf{Q}_t^{\T} (x_t^{\eta}\mathbf{I} +2 \eta^2 \mathbf{\Lambda}_t )^{-1}  \mathbf{Q}_t  \bm{v}^{\eta}_{t+1},
      & \text{otherwise,}
  \end{cases}
\]
where $\bm{v}^{\eta}_{t+1}\coloneqq \left(\mathbf{I}/D^2
+2\eta^2\mathbf{S}_t\right) \bm{u}^{\eta}_{t+1}$ and the scalar $x_{t}^{\eta}$ is the unique solution of 
\begin{align}
\rho_t^{\eta}(x) \coloneqq \sum_{i=1}^{d} \frac{\inner{\bm{e}_i}{\mathbf{Q}_t \bm{v}^{\eta}_{t+1}}^2}{(x+2\eta^2 \lambda^i_t )^2} =\frac{D^2}{4}. \label{proxyfun}
\end{align}
\end{lemma}
Since $\rho_t^{\eta}$ in \eqref{proxyfun} is strictly convex and
decreasing, $\rho_t^{\eta}(x)=D^2/4$ can be solved using Newton's method
in linear time.

\begin{algorithm}[tbp]
\begin{algorithmic}
  \REQUIRE A bounded convex set $\mathcal{U}\subset \mathbb{R}^d$ with diameter $D>0$, a Lipschitz bound $G>0$.
  \\
  We write \metagrad($D$) for \metagrad{} applied to the ball $\mathcal{B}_{D}$ enclosing $\mathcal U$.
\FOR{$t=1$ \KwTo $T$}
\STATE Get $\bh{u}_t$ from \metagrad($D$)  \tcp*{\hspace{-5mm} The initial input to \metagrad{} is $B=DG$.}
\STATE Predict $\bh{w}_t = \Pi_{\mathcal{U}}(\bh{u}_t)$ and receive $\mathring{\bm{g}}_t = \nabla \ell_t(\bh{w}_t)$;
\STATE Set $\bm{g}_t  \in \tfrac{1}{2} \left( \mathring{\bm{g}}_t +\norm{\mathring{\bm{g}}_t}  \partial \op{d}_{\mathcal{U}}(\bh{u}_t)  \right)$;
\STATE Send $\bm{g}_t$ to \metagrad($D$) ;
\ENDFOR
\caption{Reducing an OCO problem on $\mathcal{U} \subset{R}^d$ to one on a ball.}
\label{OCOGeneral}
\end{algorithmic}
\end{algorithm}
A further improvement leverages the rank-one update ${\mathbf{S}}_t = {\mathbf{S}}_{t-1}+ {\bm{g}}_t {\bm{g}}_t^{\T}$ to update $\mathbf{\Lambda}_{t-1}$
and $\mathbf{Q}_{t-1}$. It is
possible to compute the new matrices $\mathbf{\Lambda}_t$ and
$\mathbf{Q}_t$ in, respectively, $O(d^2)$ and $O(d^3)$ flops, where the
latter cost for computing $\mathbf{Q}_t$ is only due to matrix
multiplication (rather than a full singular value decomposition) \citep{bunch1978rank}, and
thus admits an efficient parallel implementation. 
 
\subsection{A Reduction to the Ball}\label{bbred.gnrlzd} In this
subsection, we extend the black-box technique of \cite{cutkosky2018} to
reduce an OCO problem on an arbitrary bounded convex set
$\mathcal{U}\subset\mathbb{R}^d$ to one on a ball, where the
implementation of \metagrad{} from the previous subsection can be applied.

Let $D$ be the diameter of a closed bounded convex set $\mathcal{U}\subset \mathbb{R}^d$ as in \eqref{rad}, so that the ball $\mathcal{B}_D$ of radius $D/2$ encloses $\mathcal{U}$. As in the previous section, we again assume, without loss of generality, that $\mathcal{U}$ is centered at $\bm{0}$. For $\bm{u}\in \mathcal{U}$, we denote $\op{d}_{\mathcal{U}}(\bm{u}) = \min_{\bm{w} \in \mathcal U} \norm{\bm{u}-\bm{w}}_2$ the \emph{distance function} from the set $\mathcal{U}$, and we define $\Pi_{\mathcal{U}}(\u)\coloneqq \{\w\in \mathcal{U}: \norm{\bm{w}-\u}_2 = \op{d}_{\mathcal{U}}(\u) \}$. Algorithm \ref{OCOGeneral}  reduces the OCO problem on the set $\mathcal{U}$ to one on the ball $\mathcal{B}_{D}$, where the \metagrad{} algorithm is used as a black-box to solve it. We note that Algorithm \ref{OCOGeneral} (including its \metagrad{} subroutine) only performs a single projection (applied to the output of  \metagrad{}) onto the set $\mathcal{U}$ in \emph{Euclidean distance} --- as opposed the \emph{time-varying Mahalanobis distance} \eqref{quadprog2}; the \metagrad{} subroutine only performs projections onto the ball $\mathcal{B}_D$, which can be done efficiently as described in the previous subsection.

In the next theorem, we assume that a Lipschitz bound $G>0$ is known in advance\footnote{If one uses \clipmetagrad{} or \restartgrad{} as the subroutine in Algorithm \ref{OCOGeneral} instead of \metagrad{}, then a Lipschitz bound need not be known in adavance; a version of Theorem \ref{reductionbound} with different constants would still hold in this case.}, and we let $\mathring{\tmp{R}}^{\bm{u}}_T \coloneqq \sum_{t=1}^T  \inner{\bh{w}_t - \bm{u}}{\mathring{\grad}_t}$ and $\mathring{\tmp{V}}^{\bm{u}}_T \coloneqq \sum_{t=1}^T \inner{\bh{w}_t - \bm{u}}{\mathring{\grad}_t}^2$ be the pseudo-regret and `variance' corresponding to Algorithm \ref{OCOGeneral}. We now show that the (pseudo) regret guarantee of \metagrad{} readily transfers to Algorithm \ref{OCOGeneral} with almost no overhead:
\begin{theorem} 
\label{reductionbound}
Let $D>0$, and suppose that the \metagrad{}($D$) subroutine of Algorithm \ref{OCOGeneral} achieves a pseudo-regret bound of the form \begin{align} \tilde{R}_T^{\u} \leq \sqrt{V^{\bm{u}}_T  \Gamma_T} + B \Gamma_T, \text{ for all $\bm{u}\in \mathcal{B}_D $}, \nonumber \end{align} 
where $\tilde{R}_t^{\u}\coloneqq \sum_{t=1}^T \inner{\pred_t - \u }{\grad_t}$, $V_t^{\u} \coloneqq \sum_{t=1}^T \inner{\pred_t - \u }{\grad_t}^2$, and $\Gamma_T =O(d\ln (T/d))$. Then, Algorithm \ref{OCOGeneral} guarantees:
\begin{align*}
\sum_{t=1}^T  \left(\ell_t(\bh{w}_t) - \ell_t(\bm{u})\right) \leq \mathring{\tmp{R}}^{\bm{u}}_T \leq  \sqrt{\mathring{\tmp{V}}^{\bm{u}}_T {\Gamma}_T} +   4 B {\Gamma}_T,  \ \text{ for all }\u\in \mathcal{U}.
\end{align*}
\end{theorem}
From the standard black-box reduction of \citet{cutkosky2018}, we would
obtain an unsatisfactory result in which $\mathring{\tmp{V}}^{\bm{u}}_T$
would be measured in terms of the fake gradients $\bm{g}_t$ that are
supplied internally to \metagrad($D$) instead of the actual gradients
$\mathring{\bm{g}}_t$. As this would not be sufficient to adapt to the
easiness conditions described in the introduction, the proof of
Theorem~\ref{reductionbound} involves an extra step to relate the variance term back
to the actual gradients.

\section{Conclusion}
\label{sec:conclusion}

We present algorithms that adapt to the Lipschitz constant of the loss
for OCO and experts, with hardly any overhead in terms of regret or
computation compared to their previous counterparts that had to know the
Lipschitz constant up-front. This fits into a larger picture of
understanding which types of adaptivity are possible at which price in
terms of additional regret and additional run time.

One surprising conclusion from our work is the following observation:
for OCO, \citet{CutkoskyBoahen2017Impossible} show that in general it is
not possible to be adaptive to both the Lipschitz constant and the
norm of the comparator $\|\u\|$ at the same time. Since the analogue of
$\|\u\|$ in the expert setting is the complexity measure
$\KL(\rho\|\pi)$, we might therefore conjecture that Lipschitz
adaptivity would also be incompatible with a quantile regret bound in
terms of $\KL(\rho\|\pi)$. However, our results show this conjecture to
be false: for experts there is no conflict. This holds even in cases
where the prior $\pi$ is not uniform, and our results can easily be
extended to a countably infinite number of experts where
$\KL(\rho\|\pi)$ cannot even be uniformly bounded.

A final and very interesting question is when is it possible to exploit
scenarios with large Lipschitz constants or loss ranges that occur only
very infrequently. An example of this is found in statistical learning
with heavy-tailed loss distributions. For such
scenarios, martingale methods (related to our potential functions)  suggest that it
may be necessary to replace in $f_t(\u,\eta)$ the `surrogate' negative quadratic term
that our algorithms include in the exponent by another
function appropriate for the specific distribution
\cite[Table~3]{linecrossing}. It is not currently clear what individual
sequence analogues can be obtained.

\acks{We thank the anonymous reviewers for feedback that improved the
presentation. Part of this work was performed while Zakaria Mhammedi was
conducting an internship at the Centrum Wiskunde \& Informatica (CWI).
This work was also supported by the Australian Research Council and
Data61.}

\DeclareRobustCommand{\VAN}[3]{#3} 
\bibliography{biblio}

\DeclareRobustCommand{\VAN}[3]{#2} 

\appendix

 \section{Proofs of Section \ref{Squint2}}
 \label{proofssquint}
 \begin{proof}{\textbf{of Lemma \ref{lem:pot.is.small}}}
  We proceed by induction on $T$. By definition $\Phi_0 = 0$. For $T \ge 0$, the definition \eqref{eq:sq.pot} gives
\[
  \Phi_{T+1}
  ~=~
  \underbrace{
    \sum_k \pi_k \int_0^\frac{1}{2 B_T} \frac{
      e^{\eta \scale{R}_{T}^k - \eta^2 \scale{V}_{T}^k}
      \del*{
        e^{\eta \scale{r}_{T+1}^k - \eta^2 (\scale{r}_{T+1}^k)^2}
        -
        1}}{\eta} \dif \eta
  }_{\fd Q_1}
  +
  \underbrace{
    \sum_k \pi_k \int_0^\frac{1}{2 B_T} \frac{e^{\eta \scale{R}_{T}^k - \eta^2 \scale{V}_{T}^k} -1}{\eta} \dif \eta
    }_{\fd Q_2}
  .
\]
To control the first term $Q_1$, we apply the so-called `prod bound'
$e^{x-x^2} \le 1+x$ for $x \ge -1/2$ \citep{cbms07} to $x = \eta \scale r_{T+1}^k$, which we may do as $\eta \scale r_{T+1} \ge - \frac{1}{2 B_T} B_T$. Linearity and the definition of the weights \eqref{eq:sq.weights}, yield the following upper-bound on the term $Q_1$
\[
 \sum_k \pi_k \int_0^\frac{1}{2 B_T} \frac{
    e^{\eta \scale{R}_{T}^k - \eta^2 \scale{V}_{T}^k}
    \eta \scale{r}_{T+1}^k
  }{\eta} \dif \eta
  ~=~
  \tuple*{
  \sum_k \pi_k \int_0^\frac{1}{2 B_T}
  e^{\eta \scale{R}_{T}^k - \eta^2 \scale{V}_{T}^k}
  \del*{\algp_{T+1} - \e_k}
  \dif \eta
  , \scale \vloss_{T+1}}
~=~
0
.
\]
To control the second term $Q_2$, we extend the range of the integral to find
\[
  Q_2
  ~\le~
  \sum_k \pi_k \int_0^\frac{1}{2 B_{T-1}} \frac{e^{\eta \scale{R}_{T}^k - \eta^2 \scale{V}_{T}^k} -1}{\eta} \dif \eta
  + \ln \frac{B_T}{B_{T-1}}
  ~=~
  \Phi_T + \ln \frac{B_T}{B_{T-1}}
  .
\]
\end{proof}
 
 \begin{proof}{\textbf{of Lemma \ref{lem:small.is.good}}}
  For any $\epsilon \in [0, 1/(2 B_{T-1})]$, we may split the potential \eqref{eq:sq.pot} as follows
  \[
    \Phi_T
    ~=~
    \underbrace{
      \sum_k \pi_k \int_0^\epsilon \frac{e^{\eta \scale{R}_T^k - \eta^2 \scale{V}_T^k} -1}{\eta} \dif \eta
      }_{\fd Q_1}
      +
      \underbrace{
        \sum_k \pi_k \int_\epsilon^\frac{1}{2 B_{T-1}} \frac{e^{\eta \scale{R}_T^k - \eta^2 \scale{V}_T^k} -1}{\eta} \dif \eta
      }_{\fd Q_2}
    .
  \]
  For convenience, let us introduce $\scale b_t \df \max_k \abs{\scale r_t^k} =  b_t \cdot B_{t-1}/B_t$ and abbreviate $\scale S_T \df \sum_{t=1}^T \scale b_t$.
  To bound the left term $Q_1$ from below, we use $e^x -1 \ge x$. Then combined with $\scale R_T^k \ge - \scale S_T$ and $\scale V_T^k \le \sum_{t=1}^{T-1} \scale b_t^2 \le B_{T-1} \scale S_T$ we find
  \[
    Q_1
    ~\ge~
    \sum_k \pi_k \int_0^\epsilon \scale{R}_T^k - \eta \scale{V}_T^k \dif \eta
    ~\ge~
    - \del*{\epsilon
      + \frac{\epsilon^2}{2} B_{T-1}} \scale S_T
    .
  \]
  For the right term $Q_2$, we use KL duality to find
  \begin{align*}
    Q_2
    &~=~
    \sum_k \pi_k \int_\epsilon^\frac{1}{2 B_{T-1}} \frac{e^{\eta \scale{R}_T^k - \eta^2 \scale{V}_T^k}}{\eta} \dif \eta
    +
      \ln \del*{2 B_{T-1} \epsilon},
    \\
    &~\ge~
    e^{-\KL \delcc*{\rho}{\pi}}
     \int_\epsilon^\frac{1}{2 B_{T-1}} \frac{e^{\eta \scale{R}_T^\rho - \eta^2 \scale{V}_T^\rho}}{\eta} \dif \eta
    +
    \ln \del*{2 B_{T-1} \epsilon}.
  \end{align*}
  Way pick the admissible $\epsilon = 1/(2(\scale S_T + B_{T-1}))$ for which $\del*{\epsilon + B_{T-1}\cdot  \epsilon^2/2} \scale S_T \le 1/2$ (as it is increasing in $\scale S_T \ge 0$ and decreasing in $B_{T-1} \ge 0$), and find
  \[
    \Phi_T
    ~\ge~
    e^{-\KL \delcc*{\rho}{\pi}}
    \int_\epsilon^\frac{1}{2 B_{T-1}} \frac{e^{\eta \scale{R}_T^\rho - \eta^2 \scale{V}_T^\rho}}{\eta} \dif \eta
    - \frac{1}{2}
    - \ln \del*{1+ \frac{\scale S_T}{B_{T-1}}},
  \]
  which we may reorganise to
  \[
    Q_3
    \df
    \ln
    \int_\frac{1}{2(\scale S_T + B_{T-1})}^\frac{1}{2 B_{T-1}} \frac{e^{\eta \scale{R}_T^\rho - \eta^2 \scale{V}_T^\rho}}{\eta} \dif \eta
    ~\le~
    \KL \delcc*{\rho}{\pi}
    + \ln \del*{
      \Phi_T
      + \frac{1}{2}
      + \ln \del*{1+ \frac{\scale S_T}{B_{T-1}}}
    }
    .
  \]
  The argument to bound the integral in $Q_3$ splits in 3 cases. Let us abbreviate $R \equiv \scale R_T^\rho$ and $V \equiv \scale V_T^\rho$. Let $\hat \eta = \frac{R}{2 V}$ be the maximiser of $\eta \to \eta R - \eta^2 V$.
  \begin{enumerate}
  \item
    First the important case, where $\intcc{\hat \eta - 1/\sqrt{2 V}, \hat \eta} \subseteq \intcc{ 1/(2(\scale S_T + B_{T+1})), 1/(2 B_{T-1})}$. Then
    \begin{align*}
      Q_3
      &~\ge~
      \ln
      \int_{\hat \eta - \frac{1}{\sqrt{2 V}}}^{\hat \eta} \frac{e^{\eta R - \eta^2 V}}{\eta} \dif \eta
      ~\ge~
      \ln
      \int_{\hat \eta - \frac{1}{\sqrt{2 V}}}^{\hat \eta} \frac{e^{\left(\hat \eta - \frac{1}{\sqrt{2 V}}\right) R - \left(\hat \eta - \frac{1}{\sqrt{2 V}}\right)^2 V}}{\eta} \dif \eta
      \\
      &~=~
        \left(\hat \eta - \frac{1}{\sqrt{2 V}}\right) R - \left(\hat \eta - \frac{1}{\sqrt{2 V}}\right)^2 V
        +
        \ln \ln \frac{\hat \eta}{\hat \eta - \frac{1}{\sqrt{2 V}}}
      \\
      &~=~
        \frac{R^2}{4 V}
        - \frac{1}{2}
        +
        \ln \ln \frac{1}{1 - \frac{\sqrt{2 V}}{R}}
        ~\ge~
        \frac{1}{2} \del*{\frac{R}{\sqrt{2 V}} -1}^2
    \end{align*}
    where the last inequality uses
    $\ln \ln (x/(x - 1))
      \ge
      1 - x$
      for $x \ge 1$, which can be easily verified by a one-dimensional plot. We conclude
      \[
        R
        ~\le~
        \sqrt{2 V} \del*{
          1+
          \sqrt{
            2 \KL \delcc*{\rho}{\pi}
            + 2 \ln \del*{
              \Phi_T
              + \frac{1}{2}
              + \ln\del*{1+ \frac {\scale S_T}{B_{T-1}}}
            }
          }
        }
        .
      \]
    \item
      Then in the case where
      $\hat \eta - 1/\sqrt{2 V} < 1/\scale S_T$, we have
      \[
        R
        ~<~
        \sqrt{2 V} + \frac{2 V}{\scale S_T}
        ~\le~
        \sqrt{2 V} + 2 B_{T-1},
      \]
      and we are done again.
  \item
    We come to the final case where $\hat \eta > 1/(2 B_{T-1})$, meaning that $R > V/B_{T-1}$. Here we use that for any $u \in \intcc{1/(2(\scale S_T + B_{T-1})), 1/(2 B_{T-1})}$
    \[
      Q_3
      ~\ge~
      \ln
      \int_u^\frac{1}{2 B_{T-1}} \frac{e^{u R - u^2 V}}{\eta} \dif \eta
      ~\ge~
      u R (1 - u B_{T-1})
      +
      \ln \ln \frac{1}{2 u B_{T-1}},
    \]
    and hence
    \[
      R
      \le
      \frac{
        Q_3
        - \ln \ln \frac{1}{2 u B_{T-1}}
      }{
        u (1 - u B_{T-1})
      }
      .
    \]
    Picking the feasible $u = (5 - \sqrt{5})/(10 B_{T-1})$ and using $- \ln \ln (5/(5 - \sqrt{5})) \le \ln 2$ yields
    \[
      R
      ~\le~
      5 B_{T-1}
      \del*{
        \KL \delcc*{\rho}{\pi}
        + \ln \del*{
          \Phi_T
          + \frac{1}{2}
          + \ln \del*{1+ \frac{\scale S_T}{B_{T-1}}}
        }
        + \ln 2
      }
      .
    \]
   Finally, using the fact that 
$$  \frac{\scale S_T}{B_{T-1}}=   \frac{1}{B_{T-1}} \sum_{t=1}^{T} \frac{B_{t-1}}{B_t} b_t  \leq 1+  \sum_{t=1}^{T-1} \frac{b_t}{B_t}$$
concludes the proof.
  \end{enumerate}
\end{proof}

\begin{figure}
  \centering
  \begin{tikzpicture}
    \foreach \i in {1,...,10} {
      \node[circle,fill,inner sep=.15em] (n\i) at (\i,0) {};
    }
    \node [circle,fill,inner sep=.15em] (begin) at (-1,0){};
    \node (dots) at (0,0){$\dots$};
    \node [below= .5cm of begin] {$1$};
    \node [below= .5cm of n10] {$T$};
    \node [below= .5cm of n6] {$\tau_2$};
    \node [below= .5cm of n3] {$\tau_1$};

    \draw (6.5,-1) -- (6.5,+1) node[above] {final restart};
    \draw (3.5,-1) -- (3.5,+1) node[above] {penultimate restart};

\draw [
thick,
    decoration={
        brace,
        mirror,
        raise=0.5cm
    },
    decorate
] (6.5+.05,-1) -- (10-.05,-1)
node [pos=0.5,anchor=north,yshift=-2em] {$\sqrt{V}$ bound};

\draw [
thick,
    decoration={
        brace,
        mirror,
        raise=0.5cm
    },
    decorate
] (3.5+.05,-1) -- (6.5-.05,-1)
node [pos=0.5,anchor=north,yshift=-2em] {$\sqrt{V}$ bound};

\draw [
thick,
    decoration={
        brace,
        mirror,
        raise=0.5cm
    },
    decorate
] (-1+.05,-1) -- (3.5-.05,-1)
node [pos=0.5,anchor=north,yshift=-2em] {tiny};

\draw [thick,red,->,decorate]  (6.5,-1.5) to [bend left=80] (1.25,-2.5);
\node[red] at (3.5,-3) {implies};

\end{tikzpicture}
\caption{Regret bounding strategy; most general case}\label{fig:bd.strategy}
\end{figure}
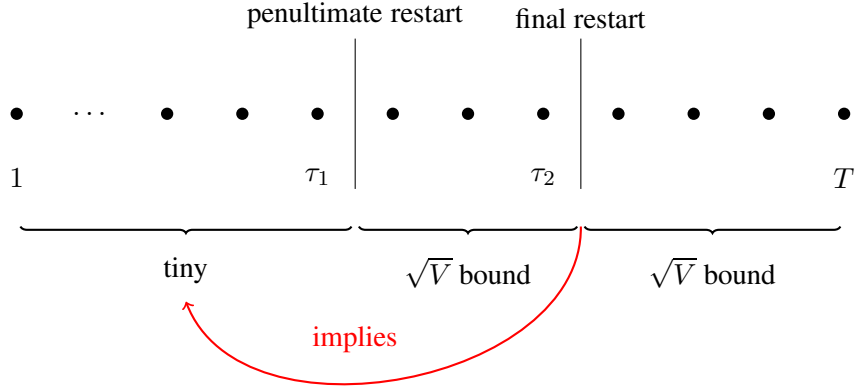

\begin{proof}{\textbf{of Theorem \ref{blackboxreduction0}}}
  The idea of the proof is to analyse the rounds in three parts, as shown in Figure~\ref{fig:bd.strategy}. 
  
  For comparator $\rho \in \triangle_K$, $B>0$ and $\tau_1,\tau_2 \in \mathbb{N}$ such that $\tau_1 < \tau_2$, we define the regret $R^{\rho}_{(\tau_1,\tau_2]}$ and variance $V^{\rho}_{(\tau_1,\tau_2]}$ of \clipsquint{} started at round $\tau_1+1$ (with input $B_{\tau_1}$) and terminated after round $\tau_2$ by
\begin{gather*}
R^{\rho}_{(\tau_1,\tau_2]}  \coloneqq \sum_{t=\tau_1+1}^{\tau_2}  \mathbb{E}_{\rho(k) } \left[r^{k}_t\right], \quad  V^{\rho}_{(\tau_1,\tau_2]}  \coloneqq \sum_{t=\tau_1+1}^{\tau_2}  \mathbb{E}_{\rho(k) } \left[(r^{k}_t)^2\right].
 \end{gather*}
 We also define
 \begin{align*}
\Gamma^{\rho}_{(\tau_1,\tau_2]} ~\df~
    \KL \delcc*{\rho}{\pi}
    + \ln \del*{
      \ln \sum_{t=1}^{\tau_2-1} \frac{b_t}{B_t}
      + \frac{1}{2}
      + \ln \left(2+  \sum_{t=\tau_1+1}^{\tau_2-1} \frac{b_{t}}{B_t} \right)
    }.
  \end{align*}
  \begin{lemma}
  \label{theclaim}
Let $\rho \in \triangle_K$ and  $\tau_1, \tau_2 \in \mathbb{N}$ be such that $\tau_1 <\tau_2$. Suppose that $B_{\tau_2-1}/B_{\tau_1} \leq \sum_{t=1}^{\tau_2-1} b_t/B_t$ (this corresponds to the case where the restart condition in line~\ref{line:runmetagrad} of Algorithm~\ref{bb1alg} is not triggered at the end of round $\tau_2-1$). Then, the regret $R^{\rho}_{(\tau_1, \tau_2]}$ of \clipsquint{} satisfies:
 \begin{gather}
R^{\rho}_{(\tau_1,\tau_2]} \leq \sqrt{2    V_{(\tau_1,\tau_2]}^\rho} \del*{
      1+
      \sqrt{2 \Gamma_{(\tau_1,\tau_2]}^{\rho}}
    }
    +
    5 B_{\tau_2}
    \del*{\Gamma_{(\tau_1,\tau_2]}^{\rho}+ \ln 2} +B_{\tau_2}. \label{eq:ziel}
 \end{gather}
 \end{lemma}
 \begin{proof}{\textbf{of Lemma~\ref{theclaim}}}
By the assumption that $B_{\tau_2-1}/B_{\tau_1} \leq \ln  \sum_{t=1}^{\tau_2-1} b_t/B_t$ and Lemma~\ref{lem:pot.is.small}, the potential function $\Phi_{\tau_2}$ can be upper-bounded by 
\begin{align*}
\Phi_{\tau_2} \leq \ln \frac{B_{\tau_2-1}}{B_{\tau_1}}\leq \ln \sum_{t=1}^{\tau_2-1} \frac{b_t}{B_t}.
\end{align*}
Using this, together with Lemma~\ref{lem:small.is.good} and \eqref{eq:ashok}, we get \eqref{eq:ziel}.
 \end{proof}
Assume without loss of generality that $b_1 \neq 0$. Then the
regret of \newsquint{} in round $t=1$ is bounded by $B_1 \leq B_T$, and \clipsquint{} is started for the first time in round $t=2$ with input $B=B_1$.

Now suppose first that the restart condition in line~\ref{line:runmetagrad} of Algorithm~\ref{bb1alg} is never
triggered, which means that $B_t/B_1\leq \sum_{s=1}^t
b_s/B_s$ for all rounds $t=2,\ldots,T$. Then for any comparator distribution $\rho \in \triangle_K$, the result follows
from Lemma~\ref{lem:small.is.good} and the facts that $V^{\rho}_{(1:T]} \leq  V_T^{\rho}$ and $\Gamma^{\rho}_{(1:T]}  \leq  \Gamma_T^{\rho}$.

Alternatively, suppose there is at least one restart. Then let $1 \leq
\tau_1 < \tau_2< T$ be such that $(\tau_1,\tau_2]$ and $(\tau_2,T]$ are
the two intervals over which the last two runs of \clipsquint{}
occurred. We invoke Lemma~\ref{lem:small.is.good} separately for both these
intervals and use Lemma~\ref{theclaim} to bound
\begin{align}
  R_{(\tau_1,T]}^{\rho}
    &\leq
      \sqrt{2    V_{(\tau_1,\tau_2]}^\rho} \del*{
      1+
      \sqrt{2 \Gamma_{(\tau_1,\tau_2]}^{\rho}}
    }
    +
    5 B_{\tau_2}
    \del*{\Gamma_{(\tau_1,\tau_2]}^{\rho}+ \ln 2} +B_{\tau_2} \nonumber\\
     &\quad+ \sqrt{2    V_{(\tau_2,T]}^\rho} \del*{
      1+
      \sqrt{2 \Gamma_{(\tau_2,T]}^{\rho}}
    }
    +
    5 B_{T}
    \del*{\Gamma_{(\tau_2,T]}^{\rho}+ \ln 2} +B_{T}, \nonumber\\
    &\leq
     2 \sqrt{    V_{(\tau_1,T]}^\rho} \del*{
      1+
      \sqrt{2 \Gamma_{(\tau_1,T]}^{\rho}}
    }
    +
    10 B_{T}
    \del*{\Gamma_{(\tau_1,T]}^{\rho}+ \ln 2} +2B_T, \label{seclast} \\
    & \leq
     2 \sqrt{    V_T^\rho} \del*{
      1+
      \sqrt{2 \Gamma_T^{\rho}}
    }
    +
    10 B_{T} 
    \del*{\Gamma_T^{\rho}+ \ln 2} +2B_T.\label{eq:onerestart}
\end{align}
where in \eqref{seclast} we used the fact that $\sqrt{x} +\sqrt{y}\leq \sqrt{2x+2y}$. If there is exactly one restart, then \eqref{eq:onerestart} implies the desired result. If there are multiple restarts, then the proof is completed by
bounding the contribution to the regret of all rounds
$2,\ldots,\tau_1$ by
\begin{align*}
  R_{(1,\tau_1]}^\u
    \leq \sum_{t=2}^{\tau_1} b_t
    \leq B_{\tau_1}\sum_{t=1}^{\tau_1} \frac{b_t}{B_t}
    \leq B_{\tau_1}\sum_{t=1}^{\tau_2} \frac{b_t}{B_t}
    < B_{\tau_2}
    \leq B_T,
\end{align*}
where the second to last inequality holds because there was a restart
at the end of round $t=\tau_2$. Finally, by bounding the instantaneous regret from the first round by $B_T$, we obtain the desired result.
\end{proof}

  \section{Proofs of Section \ref{MetaC}}
 \label{MetaGrad2proofs}
 \begin{proof}{\textbf{of Lemma \ref{lemmameta}}}
   Let $t\geq 1$. To simplify notation, we denote $\bar{r}_s^{\eta} \coloneqq \inner{\bh{u}_s - \bh{u}^{\eta}_s}{\bar{\bm{g}}_s}$, for $\bm{u}\in \mathcal{U}$ and $s\in \mathbb{N}$. By appealing to the prod-bound (\emph{i.e.} $e^{x-x^2}-1\leq x$, for $x\geq -1/2$), we have
\begin{align*}
  \Phi_{t+1}
  &~=~
    \pi(\mathcal{G}_{t+1}\setminus \mathcal{A}_{t+1})
    +  \sum_{\eta\in \mathcal{A}_{t+1}}  w^{\eta}_{t+1} \del*{e^{\eta \bar{r}^{\eta}_{t+1} - \eta (\bar{r}^{\eta}_{t+1})^2}-1}
    +  \sum_{\eta\in \mathcal{A}_{t+1}}  w^{\eta}_{t+1},
  \\
  &~\le~
    \pi(\mathcal{G}_{t+1}\setminus \mathcal{A}_{t+1})
    +  \sum_{\eta\in \mathcal{A}_{t+1}}    w^{\eta}_{t+1} \eta \bar{r}^{\eta}_{t+1}
    +  \sum_{\eta\in \mathcal{A}_{t+1}}    w^{\eta}_{t+1}.
\end{align*}
Now by \eqref{newmaster}
\[
  \sum_{\eta\in \mathcal{A}_{t+1}} w^{\eta}_{t+1} \eta \bar{r}^{\eta}_{t+1}
~=~
\sum_{\eta\in \mathcal{A}_{t+1}} \eta w^{\eta}_{t+1} (\bh{u}_{t+1} - \bh{u}^{\eta}_{t+1})^{\T} \bar{\bm{g}}_t
~=~
0.\]
Moreover, by definition of $\mathcal G_t$ and $\mathcal A_t$,
\begin{align*}
  &  \pi(\mathcal{G}_{t+1}\setminus \mathcal{A}_{t+1})
    +  \sum_{\eta\in \mathcal{A}_{t+1}}    w^{\eta}_{t+1}
  ~=~
    \pi(\set*{\eta \in \mathcal{G}_{t+1} : s_\eta > t})
    +  \sum_{\eta \in \mathcal G_{t+1} : s_\eta \le t}    w^{\eta}_{t+1},
  \\
  &~\le~
  \pi(\set*{\eta \in \mathcal{G}_{t} : s_\eta > t})
    +  \sum_{\eta \in \mathcal G_{t} : s_\eta \le t}    w^{\eta}_{t+1}
  ~=~
  \pi(\set*{\eta \in \mathcal{G}_{t} : s_\eta \ge t})
    +  \sum_{\eta \in \mathcal G_{t} : s_\eta < t}    w^{\eta}_{t+1},
  \\
  &~=~
  \pi(\mathcal{G}_{t} \setminus \mathcal A_t)
    +  \sum_{\eta \in \mathcal A_{t}}    w^{\eta}_{t+1}
    ~=~
    \Phi_t
    .
\end{align*}
Where we used that $w_{s_\eta+1}^\eta = \pi(\eta)$. Finally, as $\mathcal A_0 = \emptyset$ and $\mathcal G_0 = \mathcal G$, we find $\Phi_0 = \pi(\mathcal{G}) = 1$.
 \end{proof}
 
 \begin{proof}{\textbf{of Theorem \ref{naivemeta}}}
   Throughout this proof we will deal with slaves $\eta \in \mathcal G_T \setminus \mathcal A_T$ that are provisioned but not active yet by time $T$, and we will interpret their $s_\eta = T$ for uniform treatment, even though technically all we know from \eqref{threshold} is that $s_\eta \ge T$. 
   
First due to Lemma \ref{lemmameta}, we have $\Phi_T\leq 1$, where $\Phi_T$ is the potential defined in \eqref{masterpot}. Taking logarithms and rearranging, we find
\begin{align}
 \forall \eta \in \mathcal{G}_T,\quad  -\sum_{t=s_{\eta}+1}^T \bar{f}_t(\bh{u}^{\eta}_t,\eta)   \leq - \ln \pi(\eta).  \label{boundmaster}
\end{align}
Moreover, every slave $\eta\in \mathcal{G}_T$ guarantees the following regret for the rounds $t=s_\eta+1,\dots,T$ (see \citealt[Lemma 5]{Erven2016}):
\begin{align}
 \sum_{t=s_\eta+1}^T \left( \bar{f}_t(\bh{u}^{\eta}_t,\eta) -  \bar{f}_t(\bm{u},\eta) \right) & \leq   \ln \det\left(\mathbf{I} + 2 \eta^2 D^2 (\bar{\mathbf{S}}_T -\bar{\mathbf{S}}_{s_\eta}) \right) +\tfrac{\norm{\bm{u}}^2}{2 D^2}, \nonumber \\ &\leq  d\ln\left( 1 +\tfrac{2 D^2}{25 d B^2_{T-1}}\op{tr}{\bar{\mathbf{S}}_T} \right)+\tfrac{\norm{\bm{u}}^2}{2 D^2},\label{boundslave}
\end{align}
where in \eqref{boundslave} we used concavity of $\ln\det$, $\bar{\mathbf{S}}_{s_\eta} \succeq \bm{0}$, and the fact that $\eta \in \mathcal{G}_T \subset [0,1/(5 B_{T-1})]$.
We then invert the `wake up condition' \eqref{threshold} at time $s_\eta-1$ to infer
 \begin{align}
   -\sum_{t=1}^{s_\eta}    \bar{f}_t(\bm{u},\eta) &\le  \eta \sum_{t=1}^{s_\eta}  \bar r_t^\u
                                                      \le
                                                      \frac{
                                                      \sum_{t=1}^{s_\eta-1}   \bar r_t^\u
                                                      + \bar r_{s_\eta}^\u
                                                      }{D \sum_{t=1}^{s_\eta-1}
                                                      \norm{\bar{\bm{g}}_t}_2 + B_{s_\eta-1}}
                                                      \leq 1. \label{eq:lowbound}
 \end{align}
Combining the bounds \eqref{boundmaster}, \eqref{boundslave}, and \eqref{eq:lowbound}, then dividing through by $\eta$, gives:
\begin{align}
\forall \eta\in  \mathcal{G}_T,  \quad \cliplinregret^{\bm{u}}_T \leq \eta \clipvar^{\u}_T +\tfrac{1}{\eta} C_T(\eta),\label{gridpointregret}
\end{align}
where $C_T(\eta) \coloneqq  d\ln\left( 1 +\tfrac{2D^2}{25 d B^2_{T-1}}\op{tr}{\bar{\mathbf{S}}_T} \right) - \ln \pi(\eta)   + 2$. 

Let $C_T$ be as in the theorem statement and $\eta_*$ be the estimator defined by $\eta_*\coloneqq \sqrt{C_T/\clipvar^{\u}_T}$. Suppose that $\eta_*\leq 1/(5B_{T-1})$. By construction of the grid $\mathcal{G}_T$, there exists $i\in \mathbb{N}$ such that \begin{align}\hat\eta\coloneqq 2^{- i}/(5B_0)  \in \mathcal{G}_T \ \text{ and } \ \hat{\eta} \in \left[\eta_*/2, \eta_*\right]. \label{eq:estimator}\end{align} Since $C_T \geq 1$, the estimator $\eta_*$ can be lower-bounded by $1/\sqrt{{\clipvar}^{\bm{u}}_T}$, and thus due to \eqref{eq:estimator} we have
$2^{-i}/(5B_0) \geq 1/\sqrt{4{\clipvar}^{\bm{u}}_T}$. This implies that the prior weight on $\hat{\eta}$ satisfies
 \begin{align}
 \frac{1}{\pi(\hat{\eta})} = ( i +1)( i +2) \leq  \left( \log_2 \tfrac{2 \sqrt{ {\clipvar}^{\bm{u}}_T}}{5B_0}+1 \right) \left( \log_2 \tfrac{2 \sqrt{ {\clipvar}^{\bm{u}}_T}}{5B_0}+2\right) \leq \left(\log_2 \tfrac{ \sqrt{{\clipvar}^{\bm{u}}_T}}{B_0}+3\right)^2. \label{priorbound}
 \end{align}
Now from the fact that $1/\sqrt{{\clipvar}^{\bm{u}}_T} \leq  \eta_* \leq 1/(5B_{T-1})\leq 1/(5B_0)$, we have $\sqrt{{\clipvar}^{\bm{u}}_T} /B_0 \geq 2$. This, combined with \eqref{priorbound}, implies that $C_T(\hat{\eta})\leq C_T$, where $C_T$ is as in the theorem statement. Plugging $\eta = \hat{\eta}$ into \eqref{gridpointregret} and using the fact that $ \hat{\eta} \in\left[\eta_*/2, \eta_*\right]$, gives
\begin{align}
\label{firstbound}
\cliplinregret^{\bm{u}}_{T} \leq \hat{\eta}  \clipvar^{\bm{u}}_T + \tfrac{1}{\hat{\eta}} C_T(\hat{\eta}) \leq \eta_*  \clipvar^{\bm{u}}_T + \tfrac{2}{\eta_*} C_T =3 \sqrt{\clipvar^{\bm{u}}_T C_T}.
\end{align}
Now suppose  that $\eta_*>1/(5B_{T-1})$, and let $\hat{\eta} \coloneqq \max \mathcal{G}_T \geq 1/(10B_{T-1})$, where the last inequality follows by construction of $\mathcal{G}_T$. Note that in this case $\frac{1}{\pi(\hat{\eta})} \leq (\log_2 \frac{2B_{T-1}}{B_0}+1)(\log_2 \frac{2B_{T-1}}{B_0}+2)$, and the inequality $C_T(\hat{\eta})\leq C_T$ still holds. Plugging $\eta = \hat{\eta}$ into \eqref{gridpointregret} and using the assumption on $\eta_*$, \emph{i.e.} $\eta_*>1/(5B_{T-1})$, we obtain
\begin{align}
\label{secondbound}
\cliplinregret^{\bm{u}}_{T} \leq \hat{\eta}  \clipvar_T^{\bm{u}} + \tfrac{1}{\hat{\eta}} C_T(\hat{\eta}) \leq \hat{\eta}  \clipvar_T^{\bm{u}} + \tfrac{1}{\hat{\eta}} C_T \leq 15 B_TC_T.
\end{align}
By combining \eqref{firstbound} and \eqref{secondbound}, we get the desired result.
 \end{proof}
 
\begin{proof}{\textbf{of Theorem \ref{blackboxreduction1}}}
\label{naivemetarestartproof}
Assume without loss of generality that $b_1 \neq 0$. Then the
regret of {\restartgrad} in round one is bounded by $B_1 \leq B_T$, and
{\clipmetagrad} is started for the first time in round $t=2$ with
parameter $B=B_1$.

Let $V_{(1:T]}^\u$ and $C_{(1:T]}$ represent the quantities denoted
by $V^{\bm{u}}_T$ and $C_T$ in Theorem~\ref{naivemeta} but measured on
rounds $2,\ldots,T$. Now suppose first that the restart condition in
line~\ref{line:runmetagrad} of Algorithm~\ref{bb1alg} is never
triggered, which means that \begin{align}
\frac{B_t}{B_1}\leq \sum_{s=1}^t
\frac{b_s}{B_s}, \quad \text{ for all rounds $t=2,\dots,T$}. \label{norestart} \end{align} 
Then the result follows from Theorem~\ref{naivemeta}, $V_{(1:T]}^\u \leq V_T^\u$, for all $\u \in \mathcal{U}$, and
\begin{align}
 C_{(1:T]}
  &= d\ln\left(1 + \frac{2}{25d} \frac{\sum_{t=1}^{T-1} b_t^2 }{B^2_{T-1}}  \right) +  2 \ln \left( \log^+_2
    \frac{\sqrt{\sum_{t=2}^T b_t^2}}{B_1} +3  \right) + 2,\nonumber  \\
  &\leq d\ln\left( 1 + \frac{2}{25d} \frac{\sum_{t=1}^{T-1}
    b_t^2}{B^2_{T-1}}\right) +  2 \ln \left( \log^+_2
    \sqrt{\sum_{t=2}^T \left(\sum_{s=1}^t \frac{b_s}{B_s}\right)^2} +3 \right) + 2, \label{usethis}\\
 & \leq \Gamma_T,\nonumber
\end{align}
where in \eqref{usethis}, we used \eqref{norestart}. Alternatively, suppose there is at least one restart. Then let $1 \leq
\tau_1 < \tau_2< T$ be such that $(\tau_1,\tau_2]$ and $(\tau_2,T]$ are
the two intervals over which the last two runs of {\clipmetagrad}
occurred. We invoke Theorem~\ref{naivemeta} separately for both these
intervals to bound
\begin{align}
  R_{(\tau_1,T]}^\u
    &\leq
      3\sqrt{V_{(\tau_1,\tau_2]}^\u C_{(\tau_1,\tau_2]}}
        + 15 B_T  C_{(\tau_1,\tau_2]} + B_{\tau_2} \nonumber \\
     &\quad+ 3\sqrt{V_{(\tau_2,T]}^\u C_{(\tau_2,T]}}
        + 15 B_T  C_{(\tau_2,T]} + B_T,\nonumber \\
    &\leq
      3\sqrt{V_{(\tau_1,\tau_2]}^\u \Gamma_T/2}
      + 3\sqrt{V_{(\tau_2,T]}^\u \Gamma_T/2}
      + 15 B_T \Gamma_T + 2 B_T,  \nonumber  \\
    &\leq
      3\sqrt{V_{(\tau_1,T]}^\u \Gamma_T}
      + 15 B_T \Gamma_T + 2 B_T,\label{eq:secondstart}
\end{align}
where a subscript $(\tau_1,\tau_2]$ indicates a quantity measured only
on rounds $\tau_1+1,\ldots,\tau_2$ and the last inequality uses
$\sqrt{x} + \sqrt{y} \leq \sqrt{2x + 2y}$. If there is exactly one restart, then \eqref{eq:secondstart} implies the desired result. If there are multiple restarts, then the proof is completed by
bounding the contribution to the regret of all rounds
$2,\ldots,\tau_1$ by
\begin{align*}
  R_{(1,\tau_1]}^\u
    \leq \sum_{t=2}^{\tau_1} b_t
    \leq B_{\tau_1}\sum_{t=1}^{\tau_1} \frac{b_t}{B_t}
    \leq B_{\tau_1}\sum_{t=1}^{\tau_2} \frac{b_t}{B_t}
    < B_{\tau_2}
    \leq B_T,
\end{align*}
where the second to last inequality holds because there was a restart at $t=\tau_2$. Finally, by bounding the instantaneous regret from the first round by $B_T$, we obtain the desired result.
\end{proof}

\section{Proofs of Section \ref{four}}
\label{fourproof}
\begin{proof}{\textbf{of Lemma \ref{redquad}}}
We use the Lagrangian multiplier to solve \eqref{quadprog2}. For this, let 
\begin{align}
\mathcal{L}(\bm{u},\mu) \coloneqq \left(\bm{u}_{t+1}^{\eta}- \bm{u} \right)^{\T}\left( \mathbf{\Sigma}^{\eta}_{t+1}\right)^{-1} \left(\bm{u}_{t+1}^{\eta}- \bm{u} \right) +\mu (\bm{u}^{\T} \bm{u} -D^2).  \nonumber 
\end{align}
Setting $\frac{\partial \mathcal{L}}{\partial \bm{u}} =0$ implies that $2 \left( \mathbf{\Sigma}^{\eta}_{t+1}\right)^{-1} \left( \bm{u}-\bm{u}_{t+1}^{\eta} \right) +2\mu \bm{u}=0$. After rearranging, this becomes
\begin{align}
\bm{u} &=  \left( \left(\mu + \tfrac{1}{D^2} \right) \mathbf{I} + 2 \eta^2 {\mathbf{S}}_t    \right)^{-1}  \left(\mathbf{\Sigma}^{\eta}_{t+1}\right)^{-1} \bm{u}^{\eta}_t,\nonumber \\
& = \mathbf{Q}^{\T}_t \left(x \mathbf{I} +2 \eta^2 \mathbf{\Lambda}_t  \right)^{-1} \mathbf{Q}_t  \bm{v}^{\eta}_{t+1}, \nonumber
\end{align}
where we set $x\coloneqq \mu +1/D^2$. The result follows after observing that $\bm{u}^{\T} \bm{u} = D^2/4 \iff \rho_t^{\eta}(x)=D^2/4$. 
\end{proof}

\begin{proof}{\textbf{of Theorem \ref{reductionbound}}}
Let $\mathring{R}^{\bm{u}}_T \coloneqq \sum_{t=1}^T \inner{\bh{w}_t - \bm{u}}{\mathring{\bm{g}}_t}$ and $\mathring{V}^{\bm{u}}_T \coloneqq \sum_{t=1}^T \inner{\bh{w}_t - \bm{u}}{\mathring{\bm{g}}_t}^2$ be the pseudo-regret and `variance' of Algorithm \ref{OCOGeneral}. From our assumption on the pseudo-regret $\tilde{R}^{\bm{u}}_T$ of \metagrad{} and the fact that $2\sqrt{x} =  \inf_{\eta>0} \{\eta x + 1/\eta\}$, we have 
\begin{align}
\forall \bm{u}\in \mathcal{U}\subset \mathcal{B}_D, \forall \eta >0, \quad \eta \linregret^{\bm{u}}_T - \tfrac{\eta^2}{2} {V}^{\bm{u}}_T \leq \tfrac{1}{2}{\Gamma}_T +  \eta B \Gamma_T. \label{compbound}
\end{align} 
Now, as in the proof of \citep[Theorem~3]{cutkosky2018}, we have \begin{align}\label{cutkow} \inner{\bh{w}_t - \bm{u}}{\mathring{\bm{g}}_t} \leq 2 \mathring{\ell}_t(\bh{u}_t) - 2\mathring{\ell}_t(\bm{u}),\end{align} where $\bh{w}_t = \Pi_{\mathcal{U}}(\bh{u}_t)$ is the prediction of Algorithm \ref{OCOGeneral} at round $t$ and $\mathring{\ell}_t$ is the function defined by $\mathring{\ell}_t(\bm{u}) \coloneqq \frac{1}{2}\left( \inner{\mathring{\bm{g}}_t}{\bm{u}} + \norm{\mathring{\bm{g}}_t} \op{d}_{\mathcal{U}}(\bm{u})\right)$. By convexity of $\mathring{\ell}_t$ and the fact that ${\bm{g}}_t \in \partial \mathring{\ell}_t(\bh{u}_t)$, we have \begin{align}\inner{\bh{u}_t-\bm{u}}{{\bm{g}}_t} \geq  \mathring{\ell}_t(\bh{u}_t) - \mathring{\ell}_t(\bm{u})\geq \tfrac{1}{2}\inner{\bh{w}_t - \bm{u}}{\mathring{\bm{g}}_t}, \quad \text{for $\u \in \mathcal{U}$,} \label{useful} \end{align}
where the right-most inequality follows from \eqref{cutkow}. Since the
function $x\mapsto x - x^2/2$ is strictly increasing on the interval
$]{-\infty}, 1]$, \eqref{useful} implies that for all $\eta \in \left]0, 1/B \right] = ]0,1/(DG)]$, 
\begin{align}
 \tfrac{\eta}{2}\inner{\bh{w}_t - \bm{u}}{\mathring{\bm{g}}_t} - \tfrac{\eta^2}{8}\inner{\bh{w}_t - \bm{u}}{\mathring{\bm{g}}_t} ^2  \leq  \eta \inner{\bh{u}_t - \bm{u}}{{\bm{g}}_t} - \tfrac{\eta^2}{2} \inner{\bh{u}_t - \bm{u}}{{\bm{g}}_t}^2, \quad \text{for $\u \in \mathcal{U}$.} \nonumber
 \end{align}
Summing this over $t=1, \dots, T$ and using \eqref{compbound}, we get for all $ \eta \in \left]0, 1/B \right]$ and $\u \in \mathcal{U}$,
\begin{align}& \tfrac{1}{2} \mathring{R}^{\bm{u}}_T-\tfrac{\eta}{8} \mathring{V}^{\bm{u}}_T  \leq   \tilde{R}^{\bm{u}}_T- \tfrac{\eta}{2}{V}^{\bm{u}}_T \leq    \tfrac{1}{2\eta}{\Gamma}_T + B \Gamma_T,\quad  \text{and so} \nonumber \\
&  \mathring{R}^{\bm{u}}_T \leq  \tfrac{\eta}{4} \mathring{V}^{\bm{u}}_T+   \tfrac{1}{\eta}{\Gamma}_T +   2 B{\Gamma}_T.  \label{compbound2}
\end{align}
The `unconstrained' $\eta \in [0,+\infty]$ which minimizes the RHS of \eqref{compbound2} is given by $\eta_*\coloneqq 2\sqrt{{\Gamma}_T/\mathring{V}^{\bm{u}}_T}$. We consider two cases: suppose first that $\eta_* \leq 1/B$. For $\eta =\eta_*$, we have
\begin{align}
\label{firstbound2}
 \tfrac{\eta}{4} \mathring{V}_T^{\bm{u}} + \tfrac{1}{\eta} {\Gamma}_T =\sqrt{\mathring{V}^{\bm{u}}_T {\Gamma}_T}.
\end{align}
Now suppose that $\eta_*>1/B$. For $\eta = 1/B$, we have
\begin{align}
 \tfrac{\eta}{4} \mathring{V}_T^{\bm{u}} + \tfrac{1}{\eta} {\Gamma}_T \leq 2 B {\Gamma}_T. \label{secondbound2}
\end{align}
Combining \eqref{compbound2}--\eqref{secondbound2} yields the desired bound.
\end{proof}

\end{document}